\documentclass[11pt,a4paper]{article}
\usepackage{times,latexsym}
\usepackage{url}
\usepackage[T1]{fontenc}

\usepackage[acceptedWithA]{tacl2021v1}

\usepackage{xspace,mfirstuc,tabulary}

\newif\iftaclinstructions
\taclinstructionsfalse 
\iftaclinstructions
\renewcommand{\confidential}{}
\renewcommand{\anonsubtext}{(No author info supplied here, for consistency with
TACL-submission anonymization requirements)}
\newcommand{\instr}
\fi

\iftaclpubformat 

\else

\fi

\title{A Unifying Scheme for Extractive Content Selection Tasks}

\newcommand{\google}{$\clubsuit$}
\newcommand{\biu}{$\diamondsuit$}
\newcommand{\originai}{$\spadesuit$}

\author{Shmuel Amar\textsuperscript{\biu{}} \quad 
Ori Shapira\textsuperscript{\originai{}} \quad 
Aviv Slobodkin\textsuperscript{\biu{}\google{}} \quad
Ido Dagan\textsuperscript{\biu{}\google{}} \quad \\[1ex]
\textsuperscript{\biu{}}Bar-Ilan University \quad \textsuperscript{\google{}}Google Research \quad \textsuperscript{\originai{}}OriginAI \\[1ex]
\texttt{shmulikamar@gmail.com}
}

\usepackage{booktabs} 
\usepackage{multirow}  
\usepackage{graphicx} 
\usepackage{amsmath}  
\usepackage{svg}  
\usepackage{pifont}  
\usepackage{makecell}  
\usepackage{tikz}  
\usepackage{amssymb} 
\usepackage{tcolorbox} 
\usepackage{xcolor} 
\usepackage{relsize} 
\usepackage[normalem]{ulem} 
\usepackage{arydshln} 

\usepackage{tabularx}
\usepackage{enumitem}

\usepackage{anyfontsize}    
\newcommand{\vsmall}{\fontsize{8pt}{8pt}\selectfont}

\newif\ifshowcomments
\showcommentstrue
\showcommentsfalse

\usepackage[normalem]{ulem} 

\ifshowcomments

\usepackage{todonotes}
\usepackage{array}

\definecolor{lightgreen}{RGB}{170, 255, 200}
\definecolor{lightblue}{RGB}{179, 231, 255}
\definecolor{lightred}{RGB}{255, 179, 179}
\definecolor{lightorange}{RGB}{255, 227, 179}
\definecolor{lightpurple}{RGB}{247, 179, 255}

\newcommand{\note}[1]{\todo[inline,backgroundcolor=lightgreen]{#1}}
\newcommand{\SA}[1]{\todo[inline,backgroundcolor=lightblue]{\it SA: #1}}
\newcommand{\AS}[1]{\textcolor{lightred}{[AS] #1}}

\newcommand{\SAI}[1]{\textcolor{orange}{[SA] #1}}
\newcommand{\OS}[1]{\textcolor{cyan}{[OS] #1}}
\newcommand{\ID}[1]{\todo[inline,backgroundcolor=lightpurple]{\it ID: #1}}

\def\XXX#1{\textcolor{red}{XXX #1}}
\newcommand{\repl}[2]{%
  \if\relax\detokenize{#1}\relax
    \textcolor{blue}{#2}%
  \else
    \textcolor{red}{XXX \sout{#1}}%
    \textcolor{blue}{#2}%
  \fi
}

\else
\newcommand{\note}[1]{}
\newcommand{\SA}[1]{}
\newcommand{\SAI}[1]{}
\newcommand{\OS}[1]{}
\newcommand{\AS}[1]{}
\newcommand{\ID}[1]{}

\def\XXX#1{}
\def\repl#1#2{#2}

\fi

\newcommand{\circlet}[1]{\textcircled{\smaller{#1}}}

\newcommand{\cmark}{\ding{51}}%
\newcommand{\xmark}{\ding{55}}%

\def\ri{\textsc{GenCS}}
\def\igcsbench{\textsc{IGCS-Bench}}
\def\igcsri{\textsc{\ri{}}}
\def\igcsriunion{\textsc{\ri{}\textsubscript{Union}}}
\def\igcsrimajority{\textsc{\ri{}\textsubscript{Majority}}}

\def\riSingleStep{\textsc{\ri{}\textsubscript{1-step}}}
\def\riSingleInst{\textsc{\ri{}\textsubscript{1-inst}}}
\def\riSingleModel{\textsc{\ri{}\textsubscript{1-model}}}

\def\aspectnews{\textsc{AspectNews}}
\def\openasp{\textsc{OpenAsp}}
\def\scifact{\textsc{SciFact}}
\def\debatesum{DebateSum}
\def\spark{SPARK}

\def\aspectnewsshort{\textsc{AspSum}}
\def\debatesumshort{\textsc{ArgMine}}
\def\scifactshort{\textsc{EvidSent}}
\def\openaspshort{\textsc{AspSel}}
\def\saliencyshort{\textsc{Salience}}
\def\evidencedetectionshort{\textsc{EvidProp}}
\def\rishort{\ri{}\textsubscript{Union}}
\def\rimajorityshort{\ri{}\textsubscript{Majority}}

\def\aspectnewstiny{\makecell[cc]{\textsc{Asp} \\ \textsc{Sum}}}
\def\debatesumtiny{\makecell[cc]{\textsc{Arg} \\ \textsc{Mine}}}
\def\scifacttiny{\makecell[cc]{\textsc{Evid} \\ \textsc{Sent}}}
\def\openasptiny{\makecell[cc]{\textsc{Asp} \\ \textsc{Sel}}}
\def\saliencytiny{\makecell[cc]{\textsc{Sali-} \\ \textsc{ence}}}
\def\evidencedetectiontiny{\makecell[cc]{\textsc{Evid} \\ \textsc{Prop}}}

\def\loo{LOO}

\def\gptIcl{GPT-4\textsubscript{ICL}}
\def\gpt4{GPT-4}

\def\llama{Llama-3-8B}
\def\llamaVanilla{Llama-3-8B}

\def\llamaIcl{Llama-3-8B\textsubscript{ICL}}
\def\llamaSup{Llama-3-8B\textsubscript{Sup}}

\def\llamaRiUnion{Llama-3\textsubscript{\igcsriunion{}}}
\def\llamaRiMajority{Llama-3\textsubscript{\igcsrimajority{}}}
\def\llamaFullUnion{Llama-3\textsubscript{\loo{}+\igcsriunion{}}}
\def\llamaFullMajority{Llama-3\textsubscript{\loo{}+\igcsrimajority{}}}

\def\llamaM{Llama-3-70B}

\def\llamaL{Llama-3-405B}

\def\claude{Claude3-Opus}

\newcommand{\udash}[1]{%
    \tikz[baseline=(todotted.base)]{
        \node[inner sep=1pt,outer sep=0pt] (todotted) {#1};
        \draw[dashed] (todotted.south west) -- (todotted.south east);
    }%
}%

\newcommand{\udensdash}[1]{%
    \tikz[baseline=(todotted.base)]{
        \node[inner sep=1pt,outer sep=0pt] (todotted) {#1};
        \draw[densely dashed] (todotted.south west) -- (todotted.south east);
    }%
}%

\definecolor{myred}{RGB}{30, 136, 229} 
\definecolor{myteal}{RGB}{64, 176, 166} 
\definecolor{myorange}{RGB}{255, 193, 7} 
\definecolor{mypink}{RGB}{216, 27, 96} 

\newcommand{\colorMD}[1]{\textcolor{myred}{\udensdash{#1}}}
\newcommand{\colorGR}[1]{\textcolor{myteal}{\underline{#1}}}
\newcommand{\colorT}[1]{\textcolor{myorange}{\udash{#1}}}
\newcommand{\colorSlot}[1]{\textcolor{mypink}{\udash{#1}}}

\date{}

\begin{document}
\maketitle

\begin{abstract}
A broad range of NLP tasks involve selecting relevant text spans from given source texts. Despite this shared objective, such \textit{content selection} tasks have traditionally been studied in isolation, each with its own modeling approaches, datasets, and evaluation metrics.
In this work, we propose \textit{instruction-guided content selection (IGCS)} as a beneficial unified framework for such settings, where the task definition and any instance-specific request are encapsulated as instructions to a language model.
To promote this framework, we introduce \igcsbench{}, the first unified benchmark covering diverse content selection tasks.
Further, we create a large generic synthetic dataset that can be leveraged for diverse content selection tasks, and show that transfer learning with these datasets often boosts performance, whether dedicated training for the targeted task is available or not. Finally, we address generic inference time issues that arise in LLM-based modeling of content selection, assess a generic evaluation metric, and overall propose the utility of our resources and methods for future content selection models.\footnote{Models and datasets available at \url{https://github.com/shmuelamar/igcs}.}
\end{abstract}

\section{Introduction}

\begin{table*}[ht]
\resizebox{\textwidth}{!}{%
\begin{tabular}{@{}ll@{}}
\toprule
\textbf{Task}                      & \textbf{Instruction}                                                                                               \\ \midrule
\makecell[cl]{\textbf{\scifactshort{}}  \\\cite{wadden_fact_2020} }          & \makecell[cl]{Given the following \colorMD{abstract document(s)} of medical papers, \colorGR{select the sentences} that provide either \\ \colorT{supporting or refuting evidence for the claim:} "\colorSlot{\circlet{1} Dexamethasone decreases risk of postoperative bleeding}".}               \\[0.35cm]
\makecell[cl]{\textbf{\evidencedetectionshort{}} \\ \cite{ernst2024power}} & \makecell[cl]{Given the following \colorMD{documents} on the same topic, \colorGR{extract short and concise text phrases} that provide \\ \colorT{references to the following statement}: "\colorSlot{the Cubs put him back in the lineup}".}                                          \\[0.35cm]
\makecell[cl]{\textbf{\saliencyshort{}} \\ \cite{ernst2024power}}          & Given the following \colorMD{documents} on the same topic, \colorGR{extract short and concise} \colorT{salient} \colorGR{text phrases}.                                                                                                                                                       \\ 
\makecell[cl]{\textbf{\openaspshort{}} \\ \cite{amar-etal-2023-openasp}} & \makecell[cl]{Given the following \colorMD{news articles} on the topic "\colorSlot{\circlet{1} Hurricane Andrew}", \colorGR{extract all sentences} \\ \colorT{related to} "\colorSlot{\circlet{2} Government response}".}                                                                   \\[0.35cm]
\makecell[cl]{\textbf{\aspectnewsshort{}} \\ \cite{ahuja_aspectnews_2022}} & \makecell[cl]{Given the following \colorMD{news article}, \colorGR{select at least 1 and at most 3 sentences} \colorT{that are the most relevant to} the \\ \colorSlot{\circlet{1} Fraud's nature of the fraud: the amount of money taken, benefits for the fraudster, and how the fraud worked}.} \\[0.35cm]
\makecell[cl]{\textbf{\debatesumshort{}} \\ \cite{roush_debatesum_2020}} & \makecell[cl]{Given the following \colorMD{document}, \colorGR{select short and concise text phrases} \colorT{that summarize all the evidence} \\ \colorT{for the argument}: "\colorSlot{\circlet{1} School choice is politically popular even among dems}".}                                                \\[0.35cm] \bottomrule
\end{tabular}%
}
\caption{
Illustration of the natural language content selection instructions for the six \igcsbench{} tasks (\S\ref{subsection:bench_tasks}).
For each instruction, we highlight the \colorMD{type of single- or multi-document input}, the \colorT{requested information}, the instance-specific \colorSlot{query term(s)} (when relevant), and the \colorGR{output requirements} of the task.
}
\label{tab:igcs-instructions}
\end{table*}

Various NLP tasks essentially perform extractive \textit{content selection}, where, given a single or multiple source texts as input, the system has to select targeted spans within them as the output. 
In some tasks, particularly extractive text summarization \cite{barzilay-elhadad-1997-using, mmr}, the selection criterion is generic, where only the source texts are given as input while the output selection criteria are determined by the task itself. 
In other tasks, such as query-focused \cite{xu-lapata-2020-coarse} or aspect-based \cite{ahuja_aspectnews_2022} summarization, and evidence detection \cite{wadden_fact_2020, ernst2024power}, a specific instance-level input is provided by the user, which specifies the requested information for that instance (e.g. a query, an aspect label, or a given claim for which evidence is sought, respectively).
Additionally, content selection often becomes a natural sub-task in broader applications. For example, in attributable text generation, identifying attributions for a generated sentence in source \cite{DBLP:conf/iclr/SahaZHB23, slobodkin-etal-2024-attribute} or reference \cite{gao-etal-2023-rarr} texts is in essence a content selection task, where the instance-specific input is the model-generated sentence.
\autoref{tab:igcs-instructions} illustrates six existing content selection tasks (those included in our benchmark, described below).
In sync with the tasks illustrated in the table, we focus in this paper on the setting where the expected selected source content conveys propositional information (complete facts), typically of a substantial length.

Traditionally, such content selection tasks were considered and studied in isolation, with tailored models, datasets, and evaluation methods for each.
While earlier models relied on task-specific classifications over spans in the source texts \cite{devlin-etal-2019-bert, liu-lapata-2019-text}, recent approaches are typically based on large language models (LLMs), where task-specific information is provided in the LLM prompt. In this paper, we suggest that the success of the latter approaches opens up the opportunity for a unified framework that would address effectively a broad range of content selection tasks.

Specifically, we propose such a unified framework, termed \textit{instruction-guided content selection} (IGCS; \S\ref{subsec:task-definition}), where the task definition and instance-specific input (the ``query'', when relevant) are given to the model as an instruction in the prompt.
Following this scheme, we provide a first unified benchmark, \igcsbench{} (\S\ref{subsection:bench_tasks}), which we created by converting six existing datasets, for different content selection tasks, into a unified structure, while providing suitable instructions for each.
This benchmark facilitates the development and evaluation of general-purpose content selection models that can address multiple tasks.
In this context, we also propose to employ a particular existing evaluation metric as a generic metric for content selection settings (\S\ref{subsection:evaluation}), while showing that this metric correlates well with task-specific metrics employed in prior works.

Notably, in the context of developing fine-tuned (small) language models over training data, our unified approach facilitates investigating \textit{transfer learning} across datasets that were originally designed for different content selection settings.
In particular, we show that the performance of a generic content selection model on a specific task often improves when fine-tuned on data created for other tasks.
To further explore transfer learning across content selection settings, we leveraged top-performing LLMs to develop a larger \textit{synthetic dataset}, which comprises a broad range of content selection scenarios (\S\ref{sec:ri-dataset}).
Our results show that fine-tuning a generic model on this novel synthetic dataset improves performance across several tasks. These improvements are observed both when the synthetic data is used in a pure transfer setting, where targeted training data for the specific task at hand is not available, as well as when the synthetic data is used in combination with available task-specific training data (\S\ref{subsec:results-main}).

Finally, leveraging our unified framework, we investigate general inference-time issues that arise when utilizing LLMs for content selection, and propose strategies to reduce their impact --- namely document-level inference and aligning the output with the source documents (\S\ref{subsec:model-frag}).
Overall, we suggest that future research on either existing or novel content selection tasks, with or without targeted task-specific training data, would obtain significant gains by harnessing our provided datasets and methods.

In summary, our contributions include:  
(1) providing a unified scheme, benchmark suite, and evaluation measure for diverse content selection tasks (\S\ref{sec:bench-dataset});
(2) developing a synthetic training dataset that captures a diverse range of content selection scenarios (\S\ref{sec:ri-dataset});
(3) suggesting inference-time design choices when utilizing LLMs for content selection (\S\ref{sec:model});  
(4) investigating and assessing transfer learning benefits across diverse content selection datasets, while showing the benefits of our novel synthetic dataset (\S\ref{sec:experiments}).

\section{Background}\label{sec:background}

In this section we first provide an overview of the rich space of content selection settings (\S\ref{subsec:background-cs-tasks}), which motivates our work, and a short review of content selection modeling approaches (\S\ref{subsec:background-cs-models}).

\subsection{Content Selection Tasks}\label{subsec:background-cs-tasks}

Many end-tasks, as well as intermediate sub-tasks, can be considered as instances of a generic content selection setting, where spans within given source texts are extracted to satisfy an information need. A notable end-task example is extractive (generic) text summarization, where the output is typically a concatenation of selected salient source sentences \cite{barzilay-elhadad-1997-using, mmr,wong-etal-2008-extractive,Nallapati_2017_summarunner,zhang-etal-2023-extractive-summarization}.
Closely related, highlight summarization \cite{chen-bansal-2018-fast, arumae-etal-2019-towards, cho-etal-2020-better, ernst2024power} selects salient spans of arbitrary length, which are then highlighted for the user within their original context.
Similarly to these extractive end tasks, certain approaches for abstractive summarization employ salient-content selection as a first step, where the selected spans are then fed into an abstractive generation step \citep{chen-bansal-2018-fast,mao-etal-2020-multi,pilault-etal-2020-extractive, li-etal-2021-ease, ernst-etal-2022-proposition, adams-etal-2023-generating, wu-etal-2023-edu, DBLP:conf/iclr/SahaZHB23, slobodkin-etal-2024-attribute}.
Such approaches provide the advantage of easier traceability and attribution from the generated output texts back to the corresponding inputs from which they were generated.

In some extractive summarization variants, an instance-level input is provided to specify the requested output, such as an aspect label in aspect-based summarization \cite{ahuja_aspectnews_2022,Gunel_2023_strum,wang_2024_topicInjection}, a query in query-focused summarization \cite{xu-lapata-2020-coarse, DBLP:journals/corr/abs-2010-12694, hofmann-coyle-etal-2022-extractive}, or a question in long-form extractive question answering \cite{zhu-etal-2020-question, potluri-etal-2023-concise}.
Here again, explicit content selection from the sources either provides the eventual task output, in extractive settings, or supplies intermediate information that is further passed to an abstractive generation step.

Another content selection setting involves detecting supporting evidence for (pre-) given information. For example, in post-hoc attribution, evidence is sought for abstractive information that was (previously) generated by a model, such as in the RARR architecture \cite{gao-etal-2023-rarr}\repl{; in}{ or LAQuer \cite{hirsch-etal-2025-laquer}. In} fact-verification and evidence extraction, supporting (or refuting) source spans are sought for externally-given facts or claims \cite{thorne-etal-2018-fever, wadden_fact_2020, krishna-etal-2023-usb}. Relatedly, argument mining extracts spans, from source documents, that function as claims or evidence pertaining to a particular stance \citep{stab-gurevych-2014-annotating, roush_debatesum_2020, guo-etal-2023-aqe}.

Finally, we note that the scope of our content selection setting should be distinguished from two related settings, which fall outside our scope.
The first setting is passage retrieval \cite{10.1007/978-1-4471-2099-5_31, karpukhin-etal-2020-dense, DBLP:conf/nips/Thakur0RSG21}.
which is typically employed as an intermediate task that retrieves potentially relevant passage \textit{candidates} from a large text corpus.
These are then passed to a more precise selection module (e.g. a `reader') for making the correct selections, which are the target of our task setting
\cite{chen-etal-2017-reading, karpukhin-etal-2020-dense, arivazhagan-etal-2023-hybrid}.
The second setting involves the extraction of short phrasal spans, such as in extractive factoid question-answering \cite{rajpurkar-etal-2016-squad, kwiatkowski-etal-2019-natural, lewis-etal-2020-mlqa} and information extraction 
\cite{Cardie_1997, Xu2024LLMIE}.
In contrast, our setting focuses on tasks that require extracting text spans that jointly convey `open' propositional content.

\subsection{Modeling Content Selection}
\label{subsec:background-cs-models}

Several approaches were proposed in prior works for modeling content selection using instruction-tuned LLMs.
In cases where the output selection consisted of complete sentences, a typical approach is to split the text into sentences and then ask an LLM to generate the indices of the selected sentences \cite{parmar-etal-2024-towards, sahu-etal-2025-guide}.
This approach, however, is unsuitable when extracting spans of arbitrary length.
Other methods either augment the input with special labels or prompt the model to repeat the entire input with such labels \cite{mallick-etal-2023-adapting, sundar-etal-2024-major}, but this becomes prohibitively expensive as the input size grows.

We focus on another modeling approach, where an LLM is instructed to select the requested input spans and copy them verbatim when generating the output. Yet, as observed by \citet{ernst2024power}, models sometimes fail to verbatim copy the input as instructed, and instead generate outputs that deviate from the copied source. 
To address this issue, some works extended the decoder with constrained decoding techniques that ensure verbatim copying of input spans \cite{DBLP:conf/akbc/CastelREL22,slobodkin-etal-2024-attribute}. However, incorporating such extensions is not accessible when using API-based models.
We take a simpler approach that recovers the selected source spans via a fuzzy match with the model's output (\S\ref{subsec:model-grounding}). 

\section{Unified Content Selection Benchmark}
\label{sec:bench-dataset}

In this section, we introduce our unified content selection benchmark, \igcsbench{}, which was created by converting six existing datasets for different content selection tasks  (\S\ref{subsection:bench_tasks}) into our proposed general scheme (\S\ref{subsec:task-definition}).
Further, we propose adopting an existing and simple generic evaluation metric for content selection (\S\ref{subsection:evaluation}), which we later show to strongly correlate with four other metrics that were proposed for specific tasks (\S\ref{subsection:results-meta-eval}).

\begin{table*}[ht]
\resizebox{\textwidth}{!}{%
\begin{tabular}{@{}lr|llcc|rrr@{}}
\toprule
Task & \# Instances & Query & \makecell[cc]{Output \\ Granularity} & MD & $\varnothing$ & \makecell[cc]{Source Token Len. \\ (21--19389)} & \makecell[cc]{Selection  Token Len. \\ (0--5310)} & \makecell[cc]{Span  Token Len. \\ (0--718)}\\ \midrule
\scifactshort{} & 1109 & Claim & Sentence & \cmark & \cmark & 304   \includegraphics[height=\fontcharht\font`\B,width=2cm]{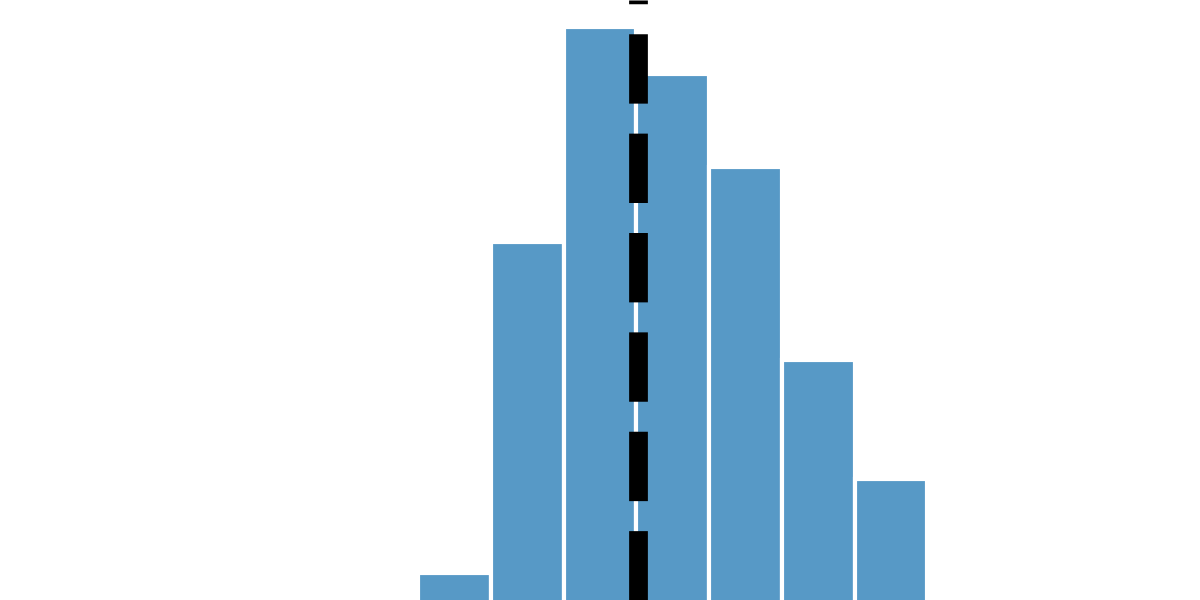}   & 46.8   \includegraphics[height=\fontcharht\font`\B,width=2cm]{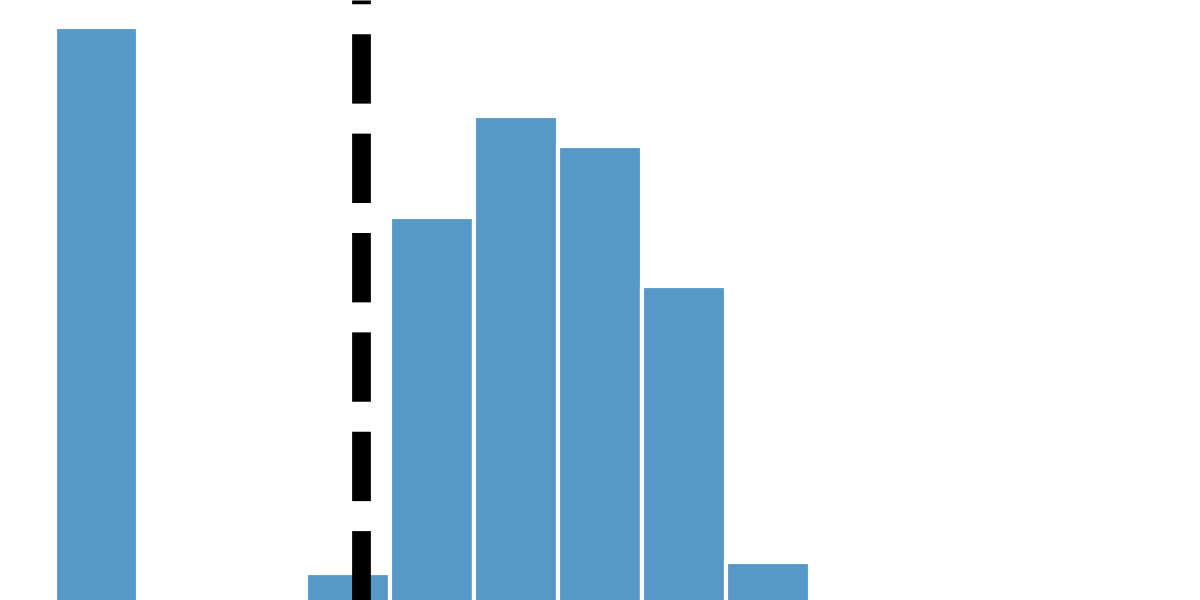}   & 23.2   \includegraphics[height=\fontcharht\font`\B,width=2cm]{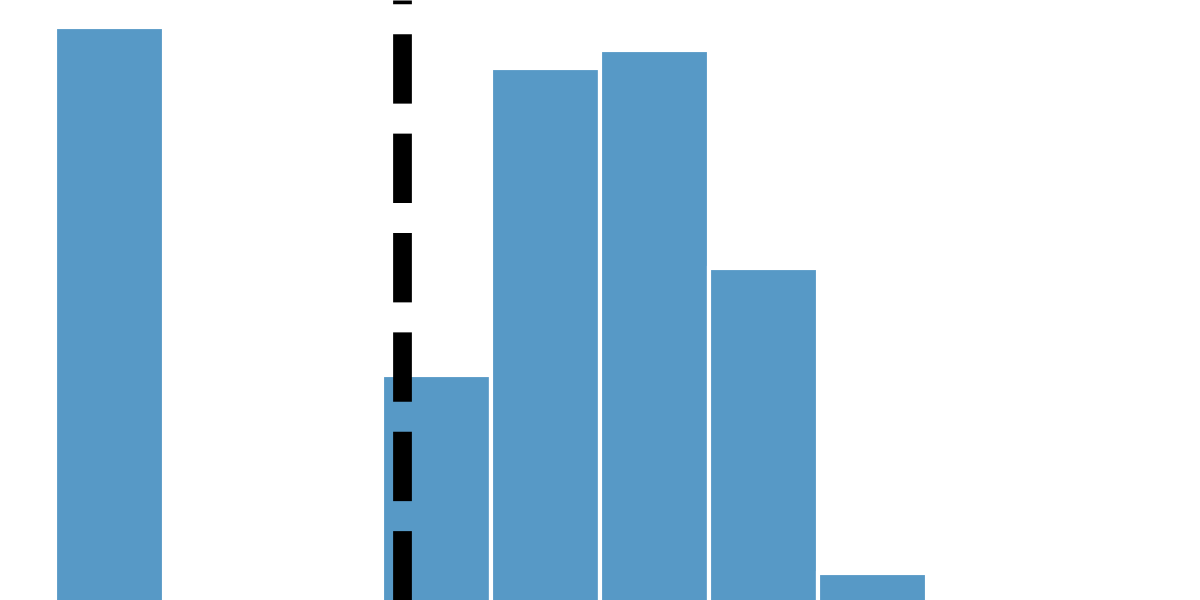}  \\
\evidencedetectionshort{} & 1332 & Proposition & Proposition & \cmark & \xmark & 2145   \includegraphics[height=\fontcharht\font`\B,width=2cm]{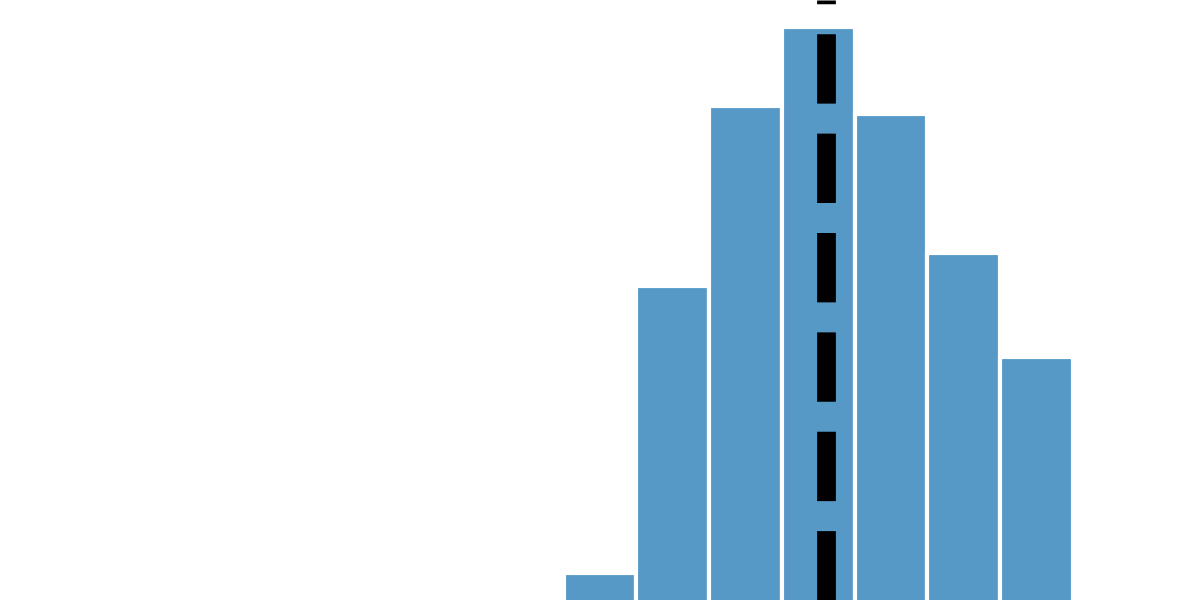}   & 32.1   \includegraphics[height=\fontcharht\font`\B,width=2cm]{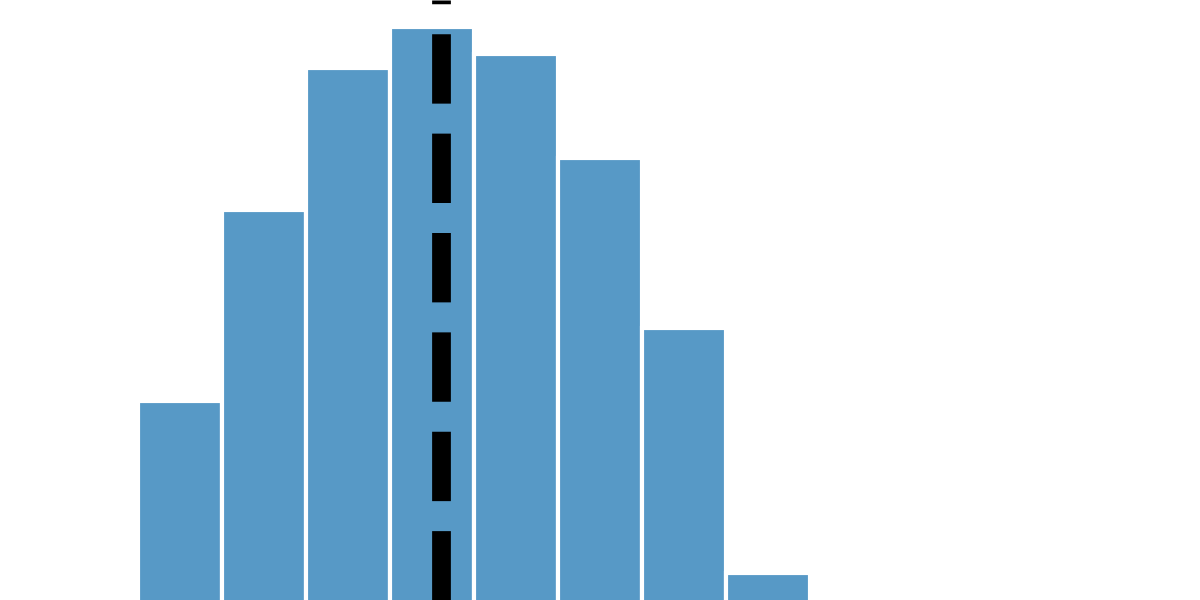}   & 14.2   \includegraphics[height=\fontcharht\font`\B,width=2cm]{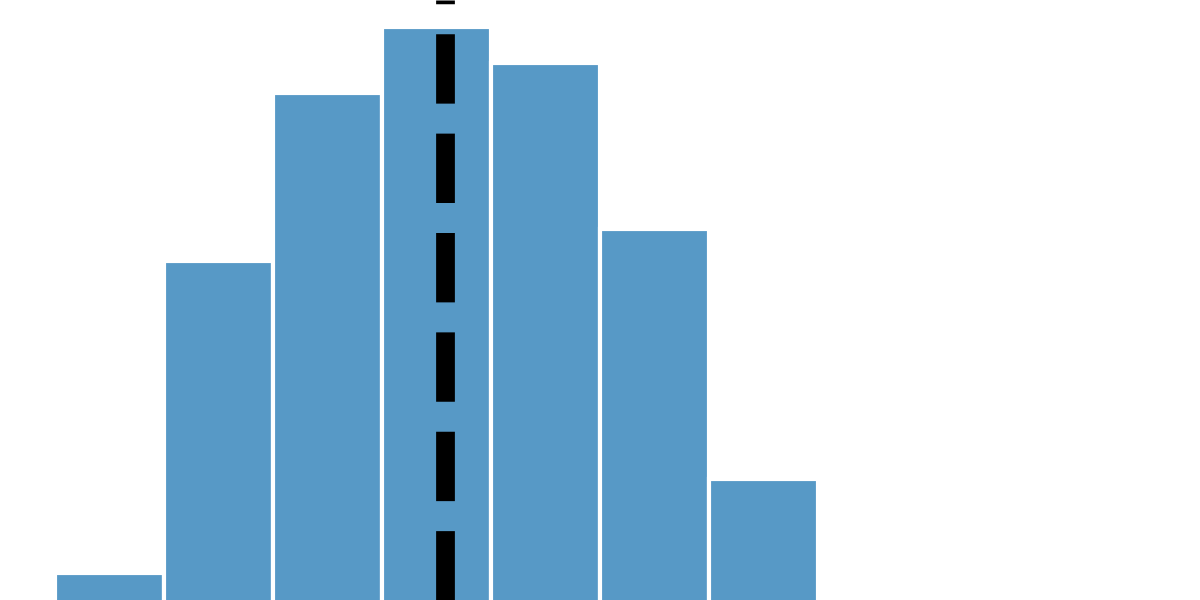} \\
\saliencyshort{} & 98 & -- & Proposition & \cmark & \xmark & 1909   \includegraphics[height=\fontcharht\font`\B,width=2cm]{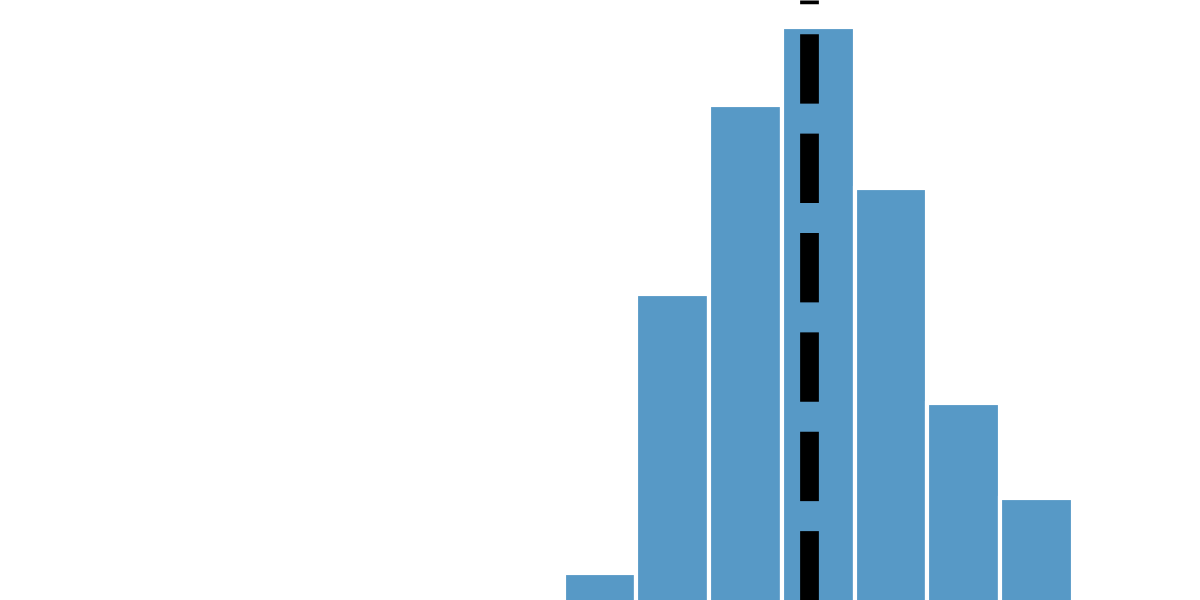}   & 436   \includegraphics[height=\fontcharht\font`\B,width=2cm]{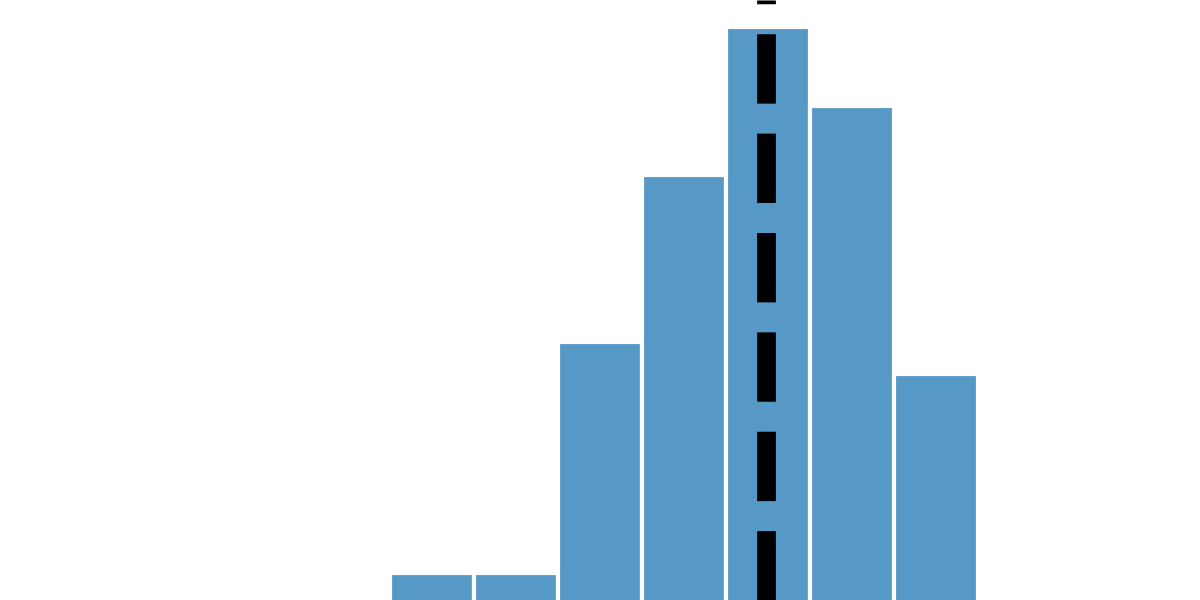}   & 12.8   \includegraphics[height=\fontcharht\font`\B,width=2cm]{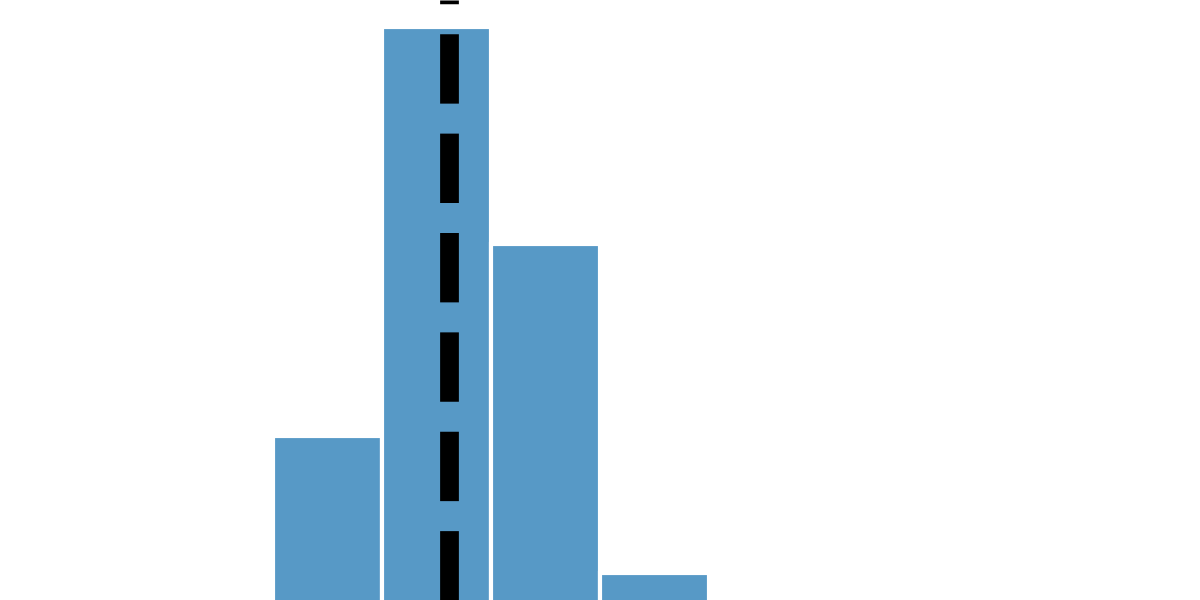} \\
\openaspshort{} & 51 & Aspect & Sentence & \cmark & \xmark & 8088   \includegraphics[height=\fontcharht\font`\B,width=2cm]{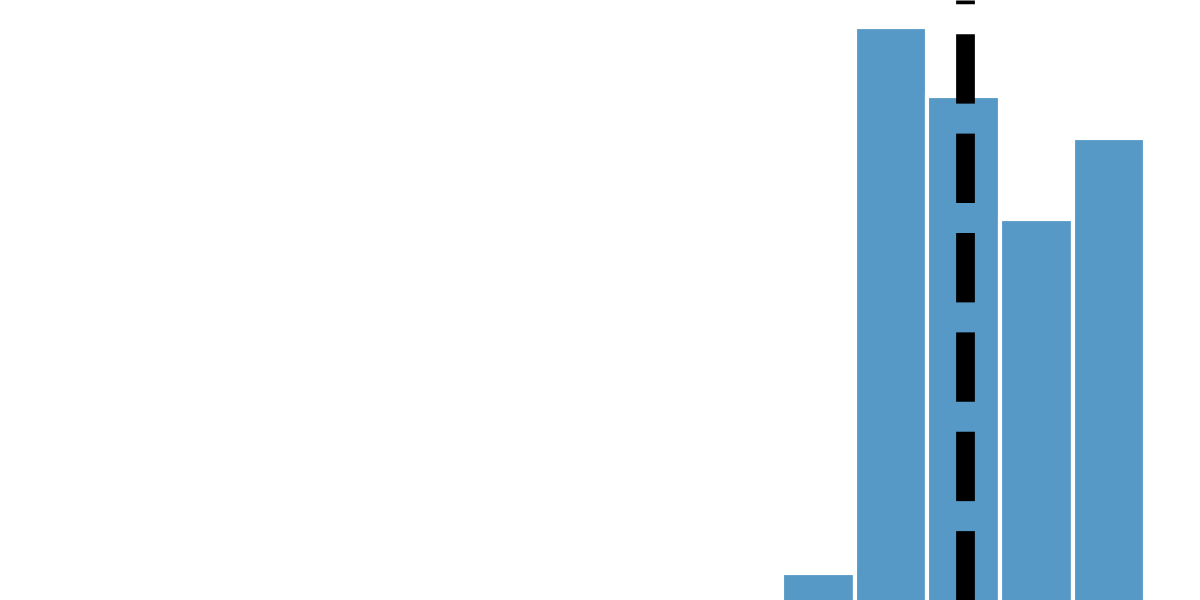}   & 955   \includegraphics[height=\fontcharht\font`\B,width=2cm]{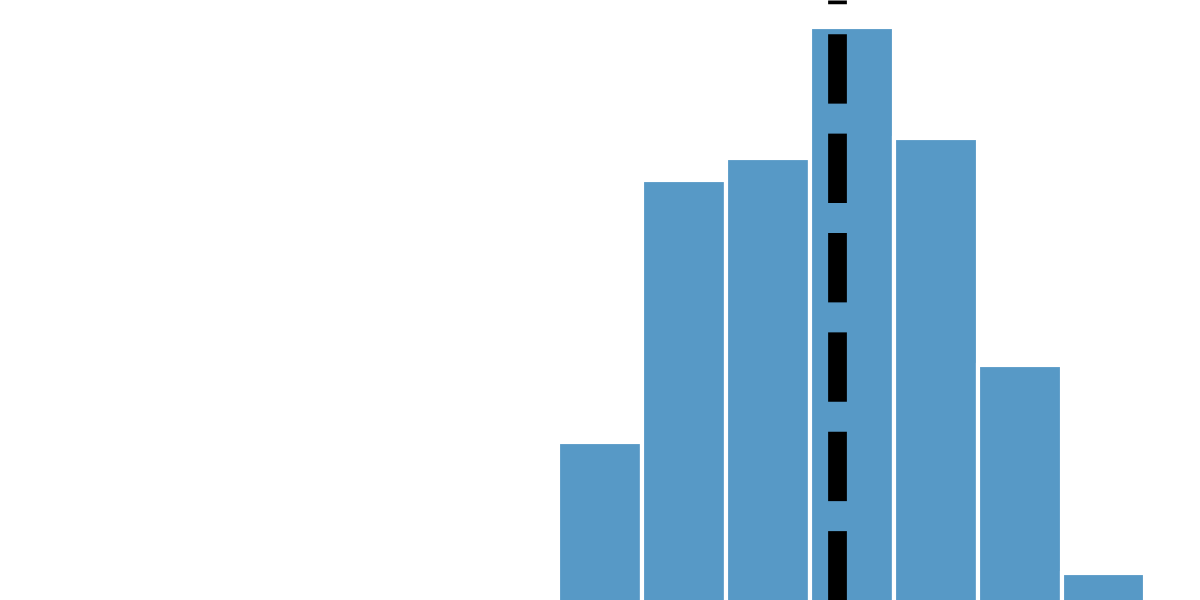} & 28.3   \includegraphics[height=\fontcharht\font`\B,width=2cm]{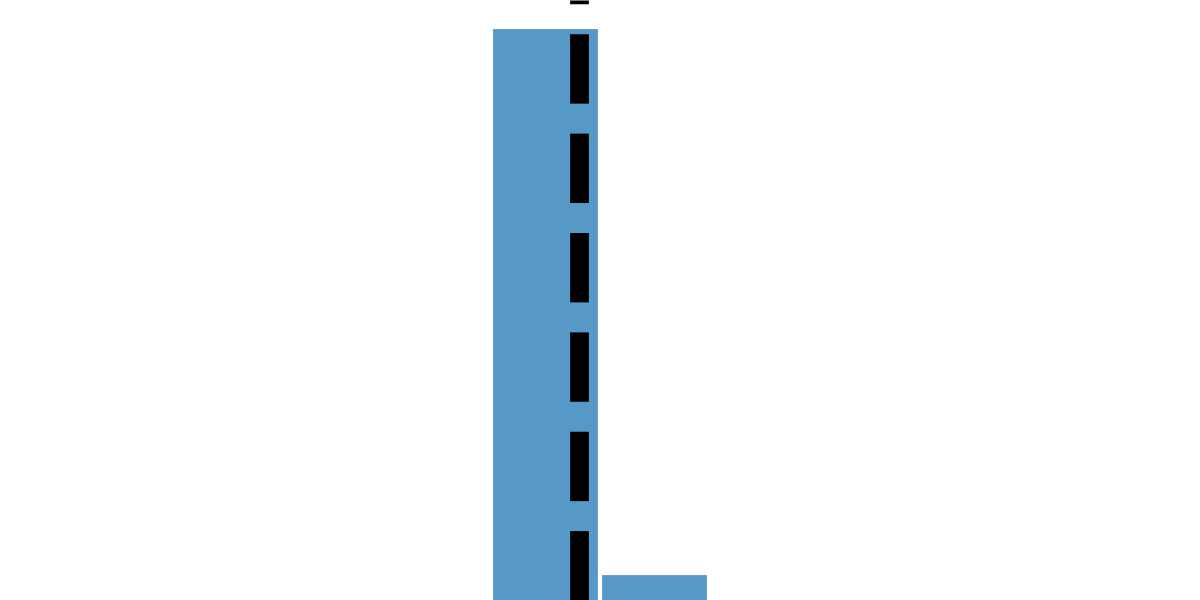}    \\
\aspectnewsshort{} & 400 & Aspect & Sentence & \xmark & \xmark & 265   \includegraphics[height=\fontcharht\font`\B,width=2cm]{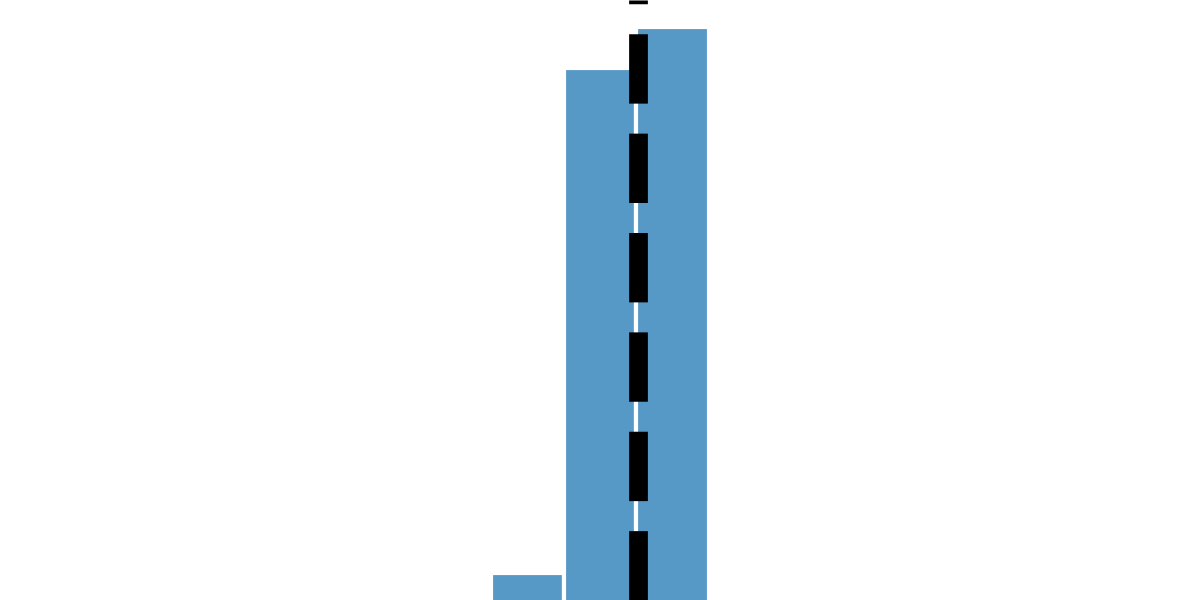}   & 83.1   \includegraphics[height=\fontcharht\font`\B,width=2cm]{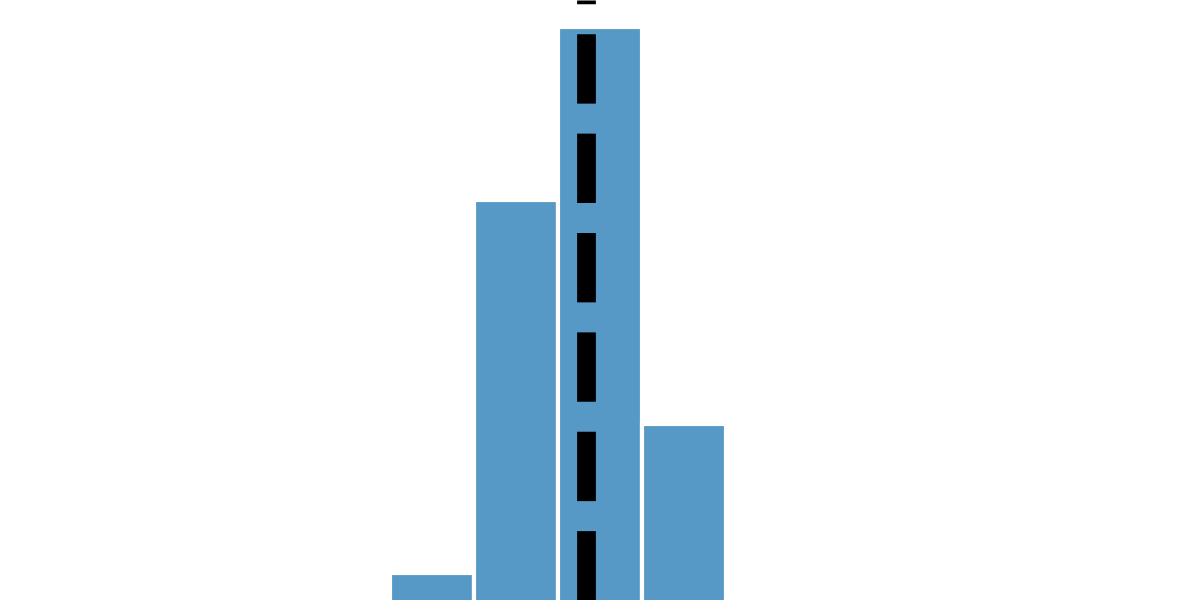}   & 30.4   \includegraphics[height=\fontcharht\font`\B,width=2cm]{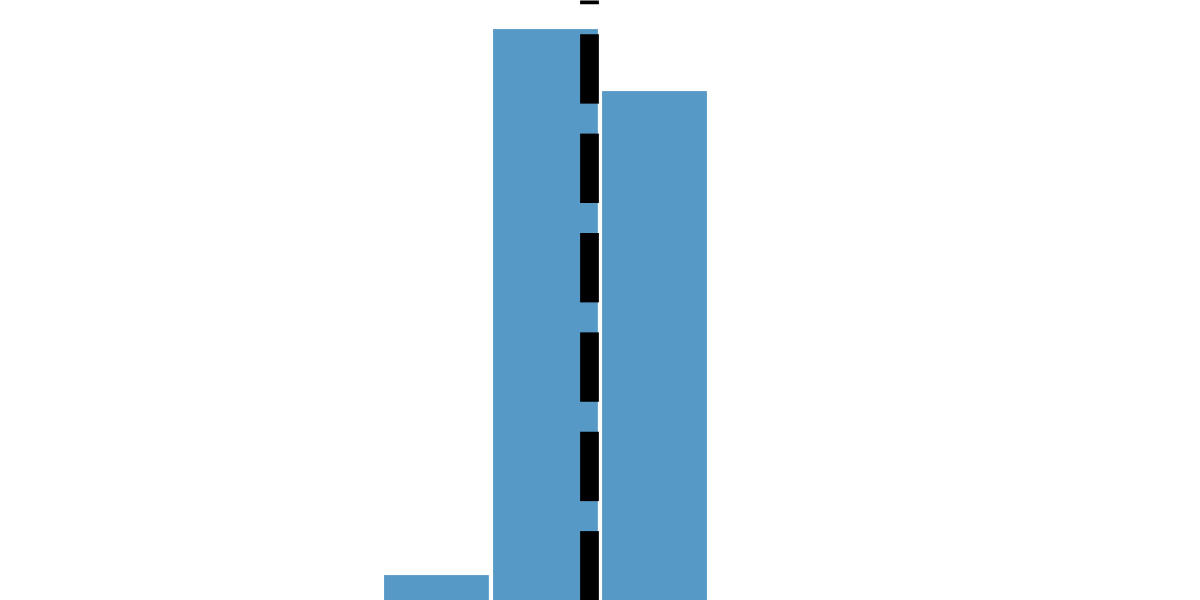} \\
\debatesumshort{} & 3000 & Argument & Span & \xmark & \xmark & 665   \includegraphics[height=\fontcharht\font`\B,width=2cm]{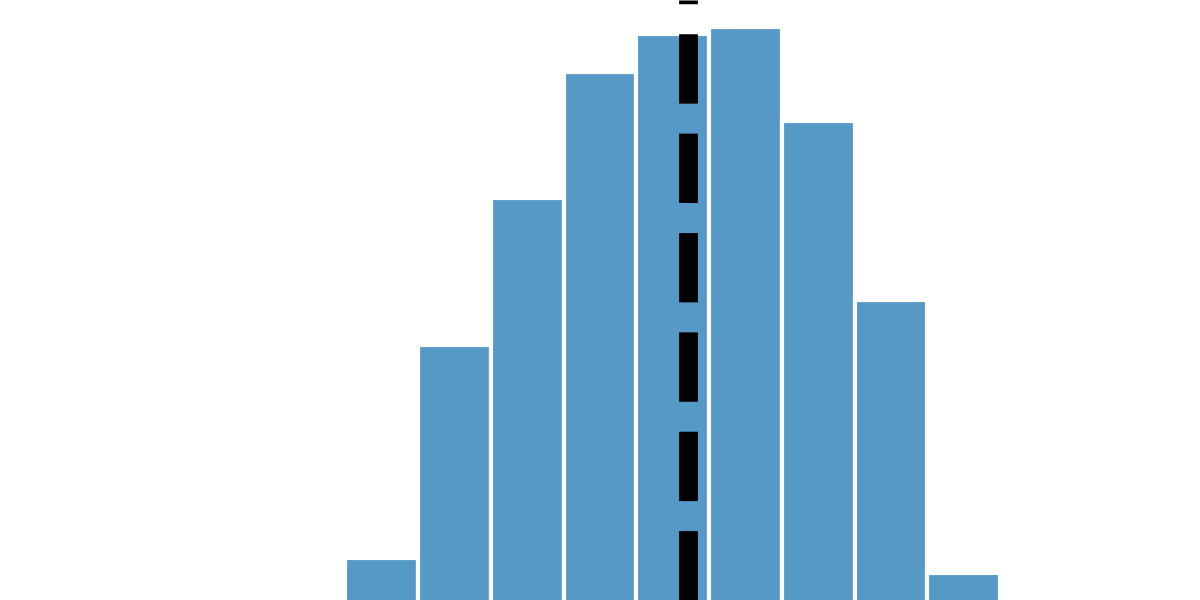}   & 244   \includegraphics[height=\fontcharht\font`\B,width=2cm]{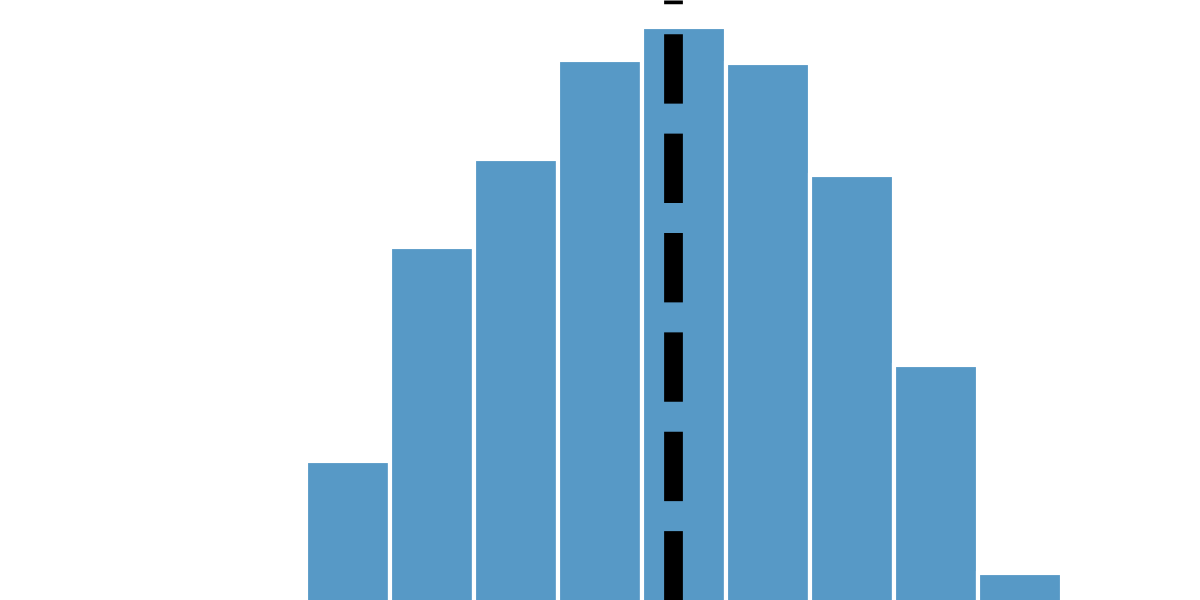}   & 25.5   \includegraphics[height=\fontcharht\font`\B,width=2cm]{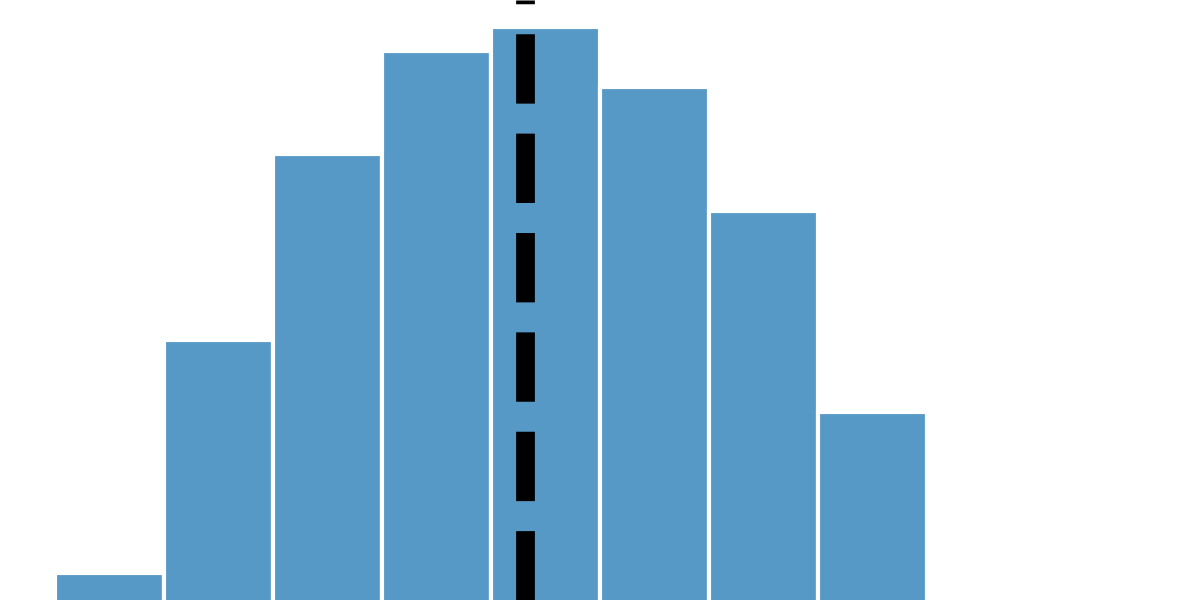} \\ \midrule
\rishort{} & 12490 & Instruction & Span & \cmark & \cmark & 1181   \includegraphics[height=\fontcharht\font`\B,width=2cm]{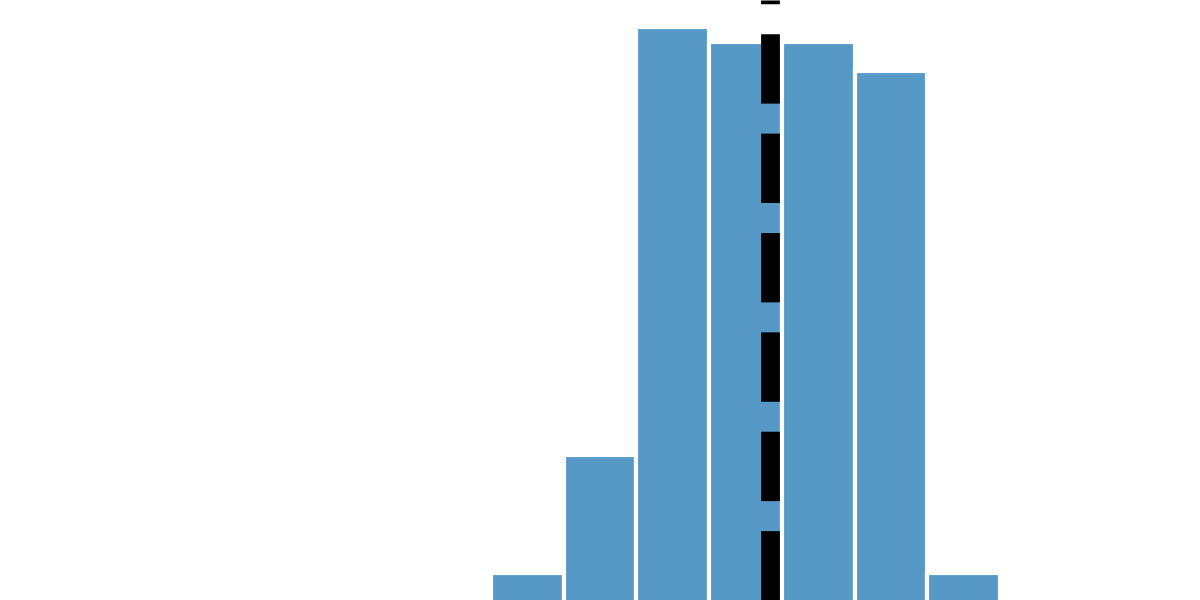}   & 86.0   \includegraphics[height=\fontcharht\font`\B,width=2cm]{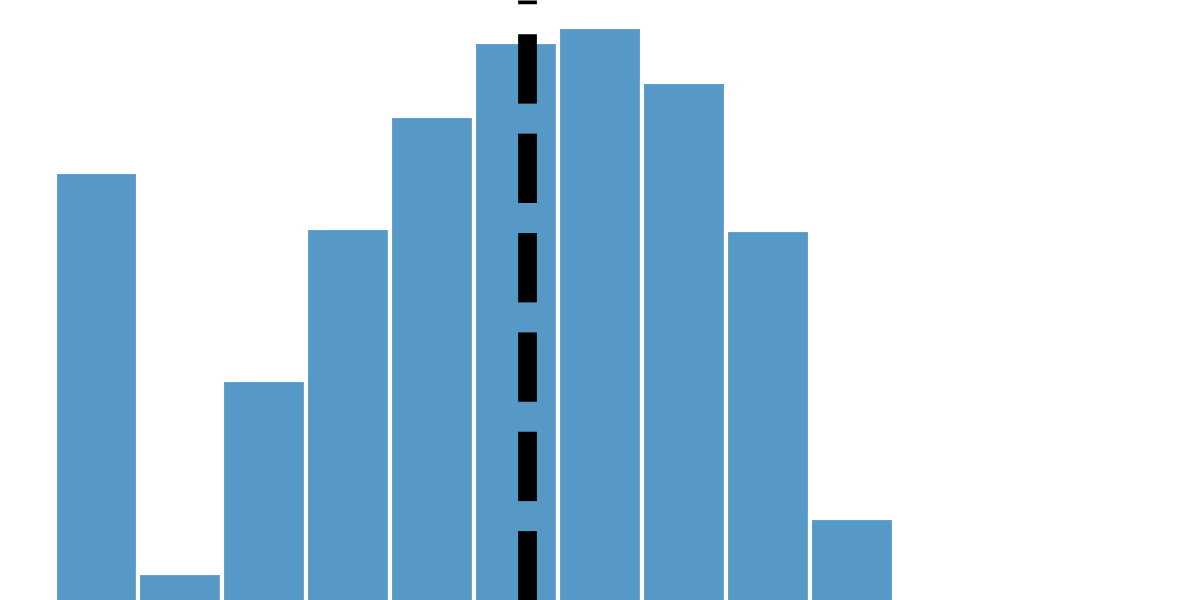}   & 30.1   \includegraphics[height=\fontcharht\font`\B,width=2cm]{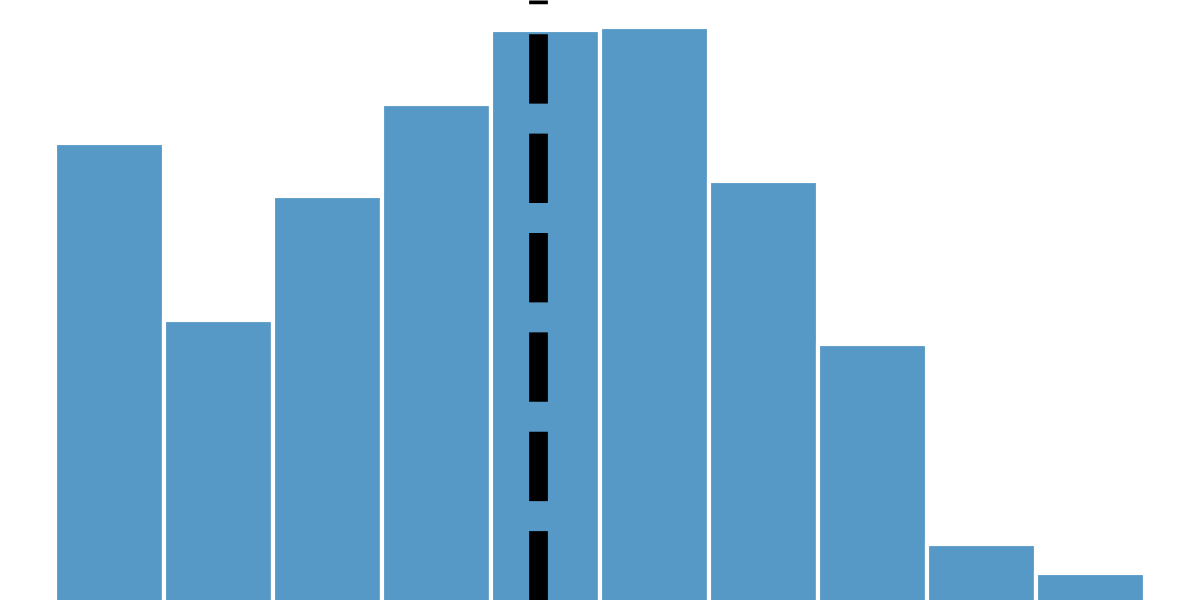} \\
\rimajorityshort{} & 12490 & Instruction & Span & \cmark & \cmark & 1181   \includegraphics[height=\fontcharht\font`\B,width=2cm]{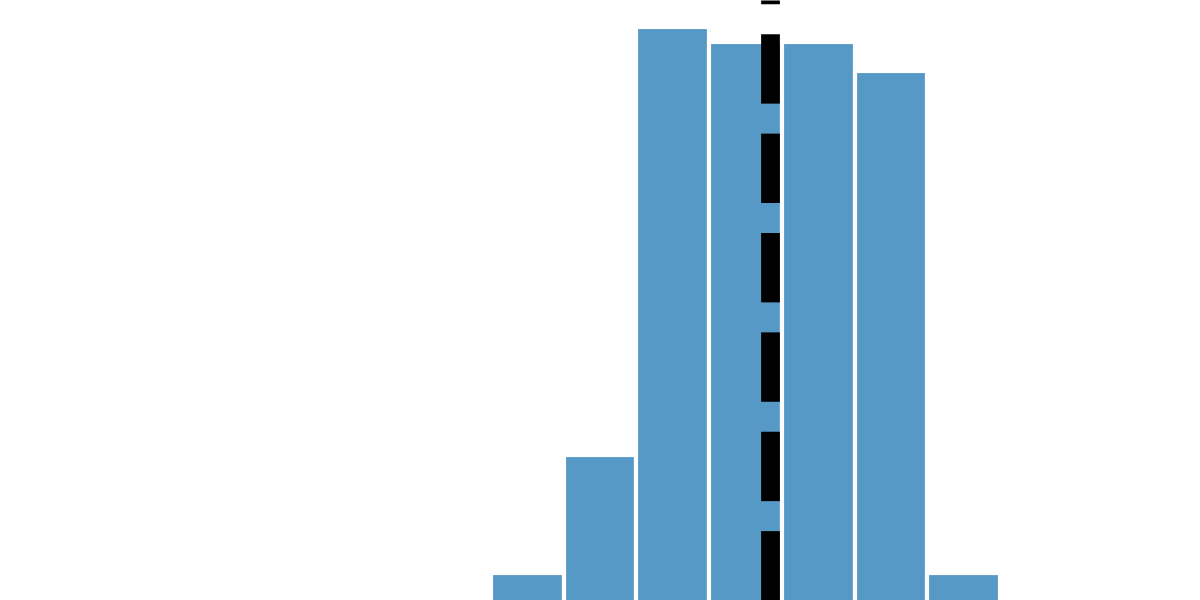}   & 75.7   \includegraphics[height=\fontcharht\font`\B,width=2cm]{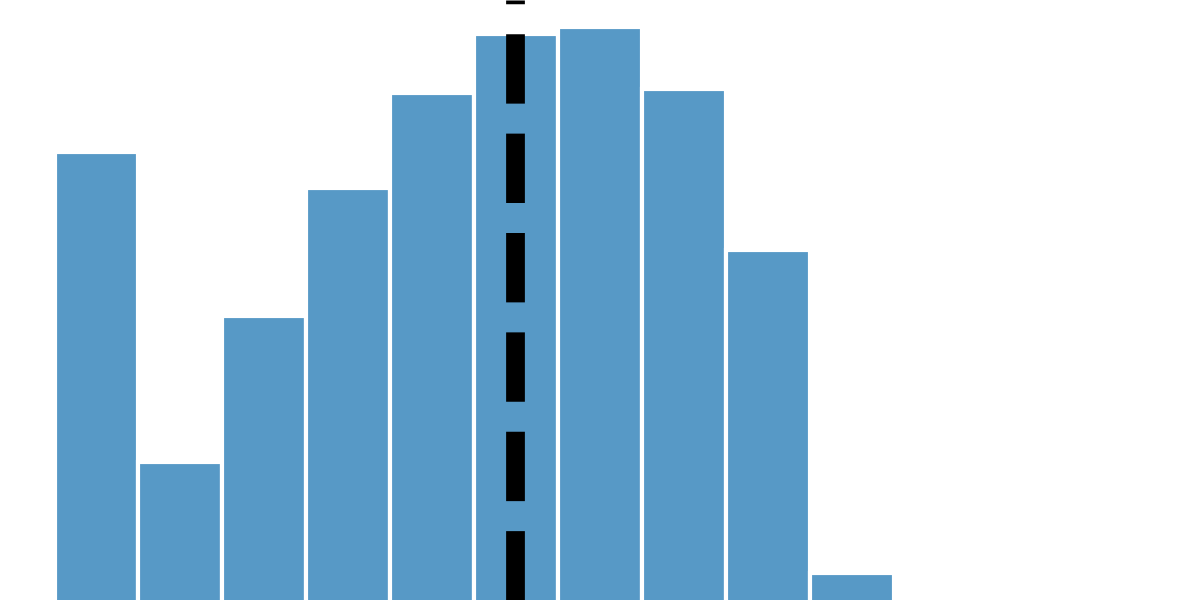}   & 28.1   \includegraphics[height=\fontcharht\font`\B,width=2cm]{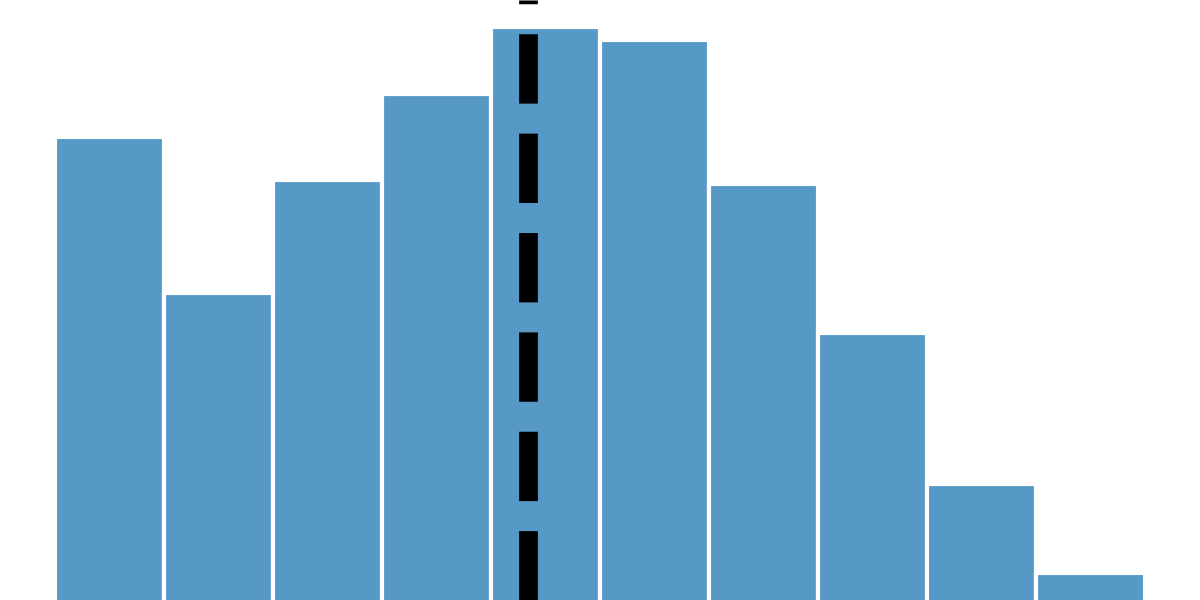} \\
\bottomrule
\end{tabular}%
}
\caption{
Content selection task properties, as detailed in \S\ref{subsec:cs-properties}. 
The six \igcsbench{} tasks (\S\ref{subsection:bench_tasks}) are at the top, and our two \igcsri{} variants (\S\ref{sec:ri-dataset}) are at the bottom. The middle section describes task setup: \textbf{Query} --- the instance-specific query type; \textbf{Output Granularity} --- the type of selections for the output; \textbf{MD} --- whether the source is multi-document or not; \pmb{$\varnothing$} --- the possibility for empty selections. The right section describes quantitative metrics, measured in token length. The histograms are shown on a log\textsubscript{2} scale on both axes, with the minimum and maximum value across all tasks displayed in the header, and the dark line marking the mean value, also written to the left of each histogram.
}
\label{tab:dataset-stats}
\end{table*}

\subsection{The IGCS Task Definition}
\label{subsec:task-definition}

A content selection task extracts a set of (typically disjoint) spans from input source texts, which jointly satisfy that specific task's information need. As mentioned earlier, our scope focuses on settings where the sought information expresses propositional information (complete facts), rather than just phrases (e.g. in response to factoid questions).
To generalize and unify the various conceivable settings of content selection, we suggest capturing the task-specific information need via 
a natural language \textit{instruction}, given as an additional input to the model, as illustrated in \autoref{tab:igcs-instructions}.

For each task, the instruction structure is defined by a pre-specified template, which may include slots that are filled with instance-specific input, termed \textit{query}.
For example, in the aspect-based sentence selection task (\openaspshort{}) \cite{amar-etal-2023-openasp}, the instruction template includes slots for the topic name associated with the input documents and the requested aspect label, pertaining to the given task instance.
The selected text spans composing the output are expected to jointly convey the overall information requested by the instruction, and to follow its output format specification, as exemplified in \autoref{tab:igcs-instructions}.

\subsection{\igcsbench{} Tasks and Datasets}
\label{subsection:bench_tasks}

To compose \igcsbench{}, we identified six existing content selection tasks with human-annotated datasets from prior works, listed below.
To ensure our benchmark data quality, we chose content selection tasks with high-quality datasets, which were created via reliable human annotation. Further, to make our benchmark useful for current research, we chose datasets on which current performance leaves room for improvement. \autoref{tab:bench-tasks-desiderata} in \autoref{app:bench-dataset-details} presents inter-annotator agreement figures and reported performances for the selected datasets.

In the original tasks, the input for each instance consists of a document set (possibly a single document), and in most cases also an instance-specific query that specifies the requested output for that instance (e.g. a specific aspect label or claim). 
To convert these tasks' instances into IGCS instances, we formulate an instruction template for each task (\autoref{tab:igcs-instructions}).
Technical details for reproducing \igcsbench{} are in \autoref{app:bench-dataset-details}.

\paragraph{Evidence Retrieval (\scifactshort{}).}
\scifact{} \cite{wadden_fact_2020} defines a task in which, given a set of medical abstracts and a scientific claim, the goal is to select sentences that either refute or support the claim.

\paragraph{Proposition-level Evidence Detection (\evidencedetectionshort{}).}
\citet{ernst2024power} define a task in which, given multiple news documents and a proposition-level text span (representing a fact), the goal is to identify all proposition-level spans within the input documents that provide evidence for the given proposition.
Compared to \scifactshort{}, this task only detects supporting evidence and targets sub-sentence spans as output.

\paragraph{Salience Detection (\saliencyshort{}).} 
\citet{ernst2024power} define a salience detection task in which, given a set of input documents, the goal is to select the most salient proposition-level spans, capturing the information that could be incorporated in a generic summary. Notice that this task generically defines the requested output, without any instance-specific input.

\paragraph{Aspect-based Sentence Selection (\openaspshort{}).}  \openasp{} \cite{amar-etal-2023-openasp} defines a task in which, given a set of documents on the same topic and an aspect label, the goal is to identify all sentences related to the specified aspect label.

\paragraph{Extractive Aspect-based Summarization (\aspectnewsshort{}).} \aspectnews{} \cite{ahuja_aspectnews_2022} 
is a dataset for extractive aspect-based summarization, where the task input is a single document and an aspect label.
The task requires selecting between 1 and 3 sentences from the document that are most relevant to the aspect. In this dataset, multiple reference selections, produced by different annotators, are provided in each instance, capturing the higher level of subjectivity in this task.

\paragraph{Argument Mining (\debatesumshort{}).} 
\debatesum{} \cite{roush_debatesum_2020} is a large argument mining dataset in which, given an article and an argument, the task is to extract all evidence spans supporting the argument.
The dataset was originally annotated for evidence to be read aloud in debate competitions.

\subsection{Evaluation Method}
\label{subsection:evaluation}

In gold instances, the reference (ground-truth) output specifies the set of source spans that should be selected for that instance. 
Prior works utilized various evaluation metrics to measure the degree of overlap between the gold and predicted source spans, often using variants of an $F_1$ measure \cite{wadden_fact_2020,ahuja_aspectnews_2022,amar-etal-2023-openasp,ernst2024power}.
As a generic evaluation metric, we suggest adopting (the existing) token-level source index comparison between the gold and predicted source spans, measuring token-level recall, precision, and $F_1$ \cite{tjong-kim-sang-buchholz-2000-introduction, ernst2024power}. As shown in \S\ref{subsection:results-meta-eval}, this metric highly correlates with the other four  evaluation metrics that were originally used for the tasks in our benchmark.

Formally, given a reference selection $S_r$ and a predicted selection $S_p$, we define $T_r$ and $T_p$ as the sets of token indices corresponding to the spans in $S_r$ and $S_p$, respectively.
We then compute precision, recall, and $F_1$ scores between $T_p$ and $T_r$.
In cases where multiple alternative reference selections are provided in the dataset, we evaluate against each reference separately, and report the scores for the reference that yields the highest $F_1$ score.\footnote{When there are multiple references, the predicted selection is expected to match just one reference, hence the maximum score amongst the references is used.}
Finally, the system level scores are reported with average $F_1$, recall, and precision across all test instances.

\paragraph{Overall scores.}\label{subsec:overall-scores}
As customary among other popular benchmark suites \citep[e.g.,][]{hendrycks2021measuring, suzgun-etal-2023-challenging}, we compute a combined score for the \igcsbench{} benchmark as the average of the individual task scores.
A combined score enables convenient comparison between models tested on the benchmark.
Specifically, each of the overall precision, recall, and $F_1$ scores is a macro-average of the corresponding token-level metrics across the six \igcsbench{} tasks.
The confidence intervals for the overall $F_1$ score is calculated with bootstrap resampling \citep[][see \autoref{app:bootstrap-overall-score} for details]{efron1994bootstrap}.
Aside from the combined token-level scores, we report each individual task's score using its original metric, as a reference point when examining a task independently. An overall ``original'' score is also computed, as the average original score across the six tasks (see \autoref{app:bench-details} for details).\footnote{As shown in \S{}\ref{subsection:results-meta-eval}, the two overall score variants correlate almost perfectly.}



\subsection{Dataset and Task Properties}
\label{subsec:cs-properties}
In the upper part of \autoref{tab:dataset-stats}, we compare notable data properties that vary across tasks in \igcsbench{}, collectively covering diverse content selection settings in our unified benchmark.

The leftmost section of the table specifies the number of instances included for each task.
Next (middle section), we compare qualitative content selection properties across the tasks.
Five of the six tasks have an instance-specific \textit{query} input, with the exception of \saliencyshort{}.
Three tasks define the \textit{output granularity} at the sentence level, while the other three operate at the sub-sentence level; for example, in \debatesumshort{}, a single-word span may be part of the larger (non-consecutive) reference selection.
Four tasks receive a multi-document set in the input (MD), which may be more challenging for a model to handle since overlapping and related content is scattered across documents.
Finally, \scifactshort{} is the only task where an empty output selection is possible ($\varnothing$), i.e., an empty set of tokens; 37\% of \scifactshort{} instances fall into this category.

Next, we compute four quantitative measurements across the six tasks (rightmost section of \autoref{tab:dataset-stats}).
The first three columns display the averages and distributions of token-level lengths of the input source text, output selection, and individual spans, for each dataset.\footnote{Throughout this paper, we use the spaCy \cite{honnibal2020spacy} tokenizer, \texttt{en\_core\_web\_sm}.}
Overall, \openaspshort{} has the largest input and output size on average, with the largest instance containing 19,389 input tokens and the largest output containing 5,310 tokens.
In \S\ref{subsec:results-inference-config}, we observe that a task's average selection size may influence modeling performance.

\section{Synthetic Dataset for IGCS}
\label{sec:ri-dataset}

\igcsbench{} (\S\ref{subsection:bench_tasks}) represents a set of prior tasks that adhere to the IGCS scheme.
Each dataset is structured differently in terms of the IGCS properties, and has a limited amount of training data.
Our objective is to facilitate large-scale fine-tuning of IGCS models over diverse sets of properties and instructions, and not only over corpora derived from particular existing tasks.
The typical approach to curate such datasets involves manual annotation which is labor-intensive.
To address this, we build upon two methods from previous works that leverage existing large-scale corpora with human-written documents.
Specifically, we employ targeted distillation, as proposed in \citet{DBLP:conf/iclr/Zhou00CP24}, to transfer knowledge from general-purpose LLMs into smaller models tailored to the specific task of content selection, by generating synthetic instructions for these documents, as done by \citet{koksal-etal-2024-longform}.

In this section we describe our three-step automated annotation pipeline which utilizes three top performing LLMs to generate two versions of a synthetic dataset for IGCS, called \textbf{Gen}eric \textbf{C}ontent \textbf{S}election (\igcsri{}) --- \igcsriunion{} and \igcsrimajority{}.
In \S\ref{subsec:gencs-ablation-results} we compare the effectiveness of several pipeline configurations for training a generic IGCS model.

\subsection{Synthetic Dataset Generation}\label{subsec:ri-dataset-generation}

An IGCS dataset includes instructions for selecting content and the corresponding selected spans from the input text sources. The process for synthesizing the dataset is as follows.

\begin{table}[t]
\resizebox{\columnwidth}{!}{%
\begin{tabular}{@{}lrrrr@{}}
\toprule
\textbf{Corpus}      & \textbf{\# Inst.} & \textbf{\# Docs} & \textbf{\makecell[cc]{Source Len.}} &  \textbf{Agr.} \\ \midrule
PubMed       & 2252                  & 1.0               & 424                   & 81.8           \\
Wikipedia    & 2025                  & 1.0               & 1428                  & 76.2           \\
Email Threads & 1991                  & 1.0               & 735                 & 69.8           \\
Books        & 1704                  & 1.0               & 1917                 & 59.2           \\
Multi-News   & 1654                  & 2.6              & 1428                 & 63.8           \\
Hotel Reviews & 1487                  & 11.0             & 1567                 & 54.4           \\
GitHub       & 1377                  & 1.9              & 1075                  & 47.2           \\ \midrule
Overall      & 12490        & 2.5     & 1181         & 65.2  \\ \bottomrule
\end{tabular}%
}
\caption{Statistics of the source corpora forming \igcsri{}.
\textbf{\# Inst.} --- total instances; \textbf{\# Docs} --- average number of documents per document set; \textbf{Source Len.} --- average tokens per document set; \textbf{Agr.} --- inter-annotator agreement among the three models on selection annotation.
}
\label{tab:ri-sources-stats}
\end{table}

\paragraph{Source Corpora.} 

Inspired by the creation of The Pile corpus \cite{DBLP:journals/corr/abs-2101-00027}, we collected single- and multi-document sets from seven corpora spanning different domains, as detailed in \autoref{tab:ri-sources-stats}.
Specifically, we leveraged news article clusters from Multi-News \cite{fabbri-etal-2019-multi}, email threads \cite{10.1007/978-3-540-30115-8_22}, English Wikipedia articles,\footnote{\url{https://en.wikipedia.org}} PubMed medical abstracts,\footnote{\url{https://huggingface.co/datasets/ncbi/pubmed}} hotel reviews \cite{10.1145/1835804.1835903}, books \cite{DBLP:conf/iclr/RaePJHL20} and GitHub code\footnote{\url{https://huggingface.co/datasets/codeparrot/github-code-clean}} (technical details in Appendix \ref{app:ri-dataset-details}).
For annotation, we sampled 500 document sets from each corpus, where each document set has an average of 1181 tokens (between 350 and 3500).\footnote{Based on \texttt{nltk.tokenize.word\_tokenize}.}

\paragraph{Step 1: Synthesizing instructions.}
In the first annotation phase we employed GPT-4\footnote{Snapshot \texttt{gpt-4-turbo-2024-04-09}.} \cite{DBLP:journals/corr/abs-2303-08774} to write five content selection instructions $I_{i}^{j}$ for every sampled document set $D_i$, encouraging generation of diverse instructions (see details and prompts in App. \ref{app:ri-dataset-prompts}).
In a real-world scenario, an instruction may yield an empty selection.
We thus asked the LLM to generate challenging instructions with no relevant content in the source text, for 5\% of the document sets in each corpus.
Overall, we gathered 17,500 instructions, i.e., 5 instructions for 500 document sets in 7 corpora.

\paragraph{Step 2: Synthesizing candidate content selections.}
In the second annotation phase, we prompted GPT-4, Claude3-Opus,\footnote{\url{https://www.anthropic.com/news/claude-3-family}, Snapshot \texttt{claude-3-opus-20240229}.} and Gemini-1.5-Pro,\footnote{\texttt{gemini-1.5-pro-latest}} to follow each of the instructions generated in the first phase and select content from the respective document set.
Since the outputs from the LLMs may deviate from the exact wording in the source documents, we aligned the outputs with the source via a grounding method, described in \S\ref{subsec:model-grounding}, and rejected spans that could not be grounded to the source text. Additionally, we discarded any instruction instance where one of the models produced a response in an invalid format.
Out of the 17,500 potential IGCS instructions, we gathered 12,490 that had three valid model selections (per corpus statistics are shown in \autoref{tab:ri-sources-stats}).

\paragraph{Step 3: Merging possible selections.}
To produce the final reference selection for an instance, we explored two natural merging strategies for the three annotated selections:
(1) the reference selection is set as the union of all selected tokens from the 3 selections, producing the recall-oriented \igcsriunion{} dataset; (2) the reference selection is the set of tokens selected by at least 2 models, producing the precision-oriented \igcsrimajority{} dataset.
As shown in \S\ref{sec:experiments}, different tasks might benefit more from either recall-oriented or precision-oriented data, hence we release both versions for future research. 
To conform to our definition of selection spans, a span is formed by concatenating consecutive selected tokens.

To conclude, this step results in the creation of the \igcsriunion{} and \igcsrimajority{} synthetic datasets. 
Each contains 12,490 instances of $(D_i, I_{i}^{j}, S_{i}^{j})$, such that the selections differ in the two datasets according to the merging strategy (in \S\ref{subsec:gencs-ablation-results} we compare different dataset generation variants).
The annotation cost is approximately \$550, with 
each dataset being twice the size of the entire \igcsbench{} and can be further expanded by annotating additional samples from the source corpora.


\subsection{\ri{} Quality and Diversity}\label{subsection:dataset-quality}

The \ri{} dataset is extrinsically evaluated in \S\ref{sec:experiments} by demonstrating its utility for transfer learning in the content selection setting.
In addition, we wish to directly assess its quality, and whether it meets our design goal of diversity in both the generated instructions and selections.
To that end, we randomly sampled from the dataset three document sets, one of which has instructions for empty selections, from each of the seven source corpora, resulting in a total of 105 dataset instances.
We instructed two annotators (NLP students) to rate each instruction and perform content selection as detailed below.

\paragraph{Quality of instructions.}
\label{subsection:dataset-quality-instructions}

A diverse dataset is expected to comprise instructions for various content selection use cases that require varying levels of informational \textit{specificity}. The instructions should also be \textit{natural} in the context of the given document set.
Accordingly, the annotators rated each instruction on a Likert scale of 1 to 5 for \textit{naturalness} and \textit{specificity} (articulated in \autoref{fig:annotation-guidelines} in App. \ref{app:ri-qa}).
The high average score of 4.0 ($\sigma=1.3$) for \textit{naturalness} indicates that, overall, the instructions are plausible and relevant to their document sets.
The average \textit{specificity} score of 3.1 ($\sigma=0.8$) indicates that the values are scattered around 3,
which reflects a broad range of scenarios, where the requested information varies from generic to anecdotal in relation to the topic of the document set.

\paragraph{Quality of selections.}

A high-quality selection must accurately adhere to the given instruction by including only the relevant text spans from the input sources.
To assess the selections in the \ri{} dataset
we measured their agreement with human-annotated selections.

We instructed human annotators to manually perform the content selection task for the sampled instances
(see the annotation interface in \autoref{fig:qa_annotation_interface} in \autoref{app:ri-qa}), and computed the inter-rater agreement between annotations, following \citet{hripcsak_agreement_2005} (see Appendix \ref{app:ri-qa-iaa} for more details).
Overall, we measured a Cohen's $\kappa$ score of 0.7 among the three models, 0.59 among the two human annotators, and 0.61 human-LLM agreement, which indicate moderate to high agreement.
In addition to preventing the significant effort from human annotators (as reported by our annotators), the LLMs were evidently capable of reliably producing selections.

Finally, selection diversity is measured through our content selection properties, as presented in the lower part of \autoref{tab:dataset-stats}.
The histograms in the table demonstrate diverse source, selection, and span sizes.
This diversity emulates a wide range of content selection scenarios, including single large-span selections, empty selections, multiple short-span selections, and selections spanning multiple text sources.

\section{Modeling}
\label{sec:model}

Following our proposed unified scheme for content selection (\S\ref{subsec:task-definition}), our main objective is to assess whether, when modeling a particular content selection task, we can leverage generic training data such as \igcsri{}. Accordingly, the focus of our modeling is to apply such transfer learning in different configurations, by fine-tuning feasibly-sized small language models, as described in \S\ref{subsec:ft-models}.\footnote{We fine-tune small models of up to 8B parameters, which fits typical research computation budgets.}
In addition, we address, at inference time, two issues that arise when applying LLMs for content selection, by fragmenting the inference to apply over one document at a time, and by post-hoc matching between the generated output and their corresponding source spans (\S\ref{subsec:models-infer-config}).


\subsection{Transfer Learning Configurations} \label{subsec:ft-models}
To model an Instruction-guided Content Selection (IGCS) task, we prompt an LLM, for each task instance, with the instance-specific instruction along with the source texts (or text) for that instance.
In order to evaluate the effects of transfer learning between different content selection tasks, we fine-tune a popular LLM, \textbf{Llama-3-8B}, with various mixtures of training data, while utilizing the training datasets in \igcsbench{} (\S{\ref{sec:bench-dataset}) and our synthetic \igcsri{} dataset (\S\ref{sec:ri-dataset}).
In our fine-tuned models, we address two transfer learning scenarios: (1) a \textit{\textbf{transfer-only}} setting, when no training data for the targeted task is available, thus fine-tuning the model only over data for other tasks; (2) \textit{\textbf{supervision+transfer}}, where training data for the targeted task is available, we use the same training data in the transfer-only setting but additionally include training data for the targeted task.
To corroborate the robustness of the observed trends and behaviors in the Llama model, we also conduct analyses using fine-tuned models from other families, namely \textbf{Qwen2.5} \cite{DBLP:journals/corr/abs-2412-15115} and \textbf{SmolLM2} \citep{DBLP:journals/corr/abs-2502-02737}. These families offer multiple small-scale models that can be fine-tuned with modest computational resources.

We test our fine-tuned models over each of the six tasks in \igcsbench{}.
For transfer-only training, we fine-tune with two different compositions of data:
(1) Leave-one-out \textbf{(\loo{})} --- mixing all available training sets in \igcsbench{} (available for \scifactshort{}, \aspectnewsshort{}, and \debatesumshort{}), except for the set of the targeted task being tested (simulating ``out-of-domain'' testing);
(2) synthetic dataset (\textbf{\ri{}}) --- fine-tuning over one of the automatically generated dataset variants, \igcsriunion{} or \igcsrimajority{}.
Analogously, in the supervision+transfer setting, we mix the training data of the targeted task with the same two compositions above of transfer data.\footnote{We note that combining the two types of transfer data did not yield notable improvements in our experiments.}
See \autoref{app:models-training-details} for technical details.


\subsection{Prompt-based Models}
\label{subsection:prompt-models}
As reference points for the results of our fine-tuned transfer models, we also report results for larger LLMs, obtained via zero- and few-shot prompting.
Specifically, for zero-shot prompting, we use the proprietary \textbf{\gpt4{}},\footnote{\texttt{gpt-4-turbo-2024-04-09}} and \textbf{\claude{}}\footnote{\texttt{claude-3-opus-20240229}} models, as well as the open-source \textbf{Llama-3} family of models \cite{DBLP:journals/corr/abs-2407-21783} --- of 8B,\footnote{\texttt{meta-llama/Meta-Llama-3-8B-Instruct}} 70B, and 405B\footnote{70B and 405B with \url{https://www.together.ai/blog/meta-llama-3-1}} parameters.
For the few-shot in-context learning setting, we experimented with \gpt4{} and Llama-3-8B \cite{dong2024surveyincontextlearning}, denoted \textbf{\gptIcl{}} and \textbf{\llamaIcl{}}, respectively.
After experimenting with the number of in-context examples, we found 2-shot to perform best.



\useunder{\uline}{\ul}{}
\begin{table*}[t]
\resizebox{\textwidth}{!}{%
\begin{tabular}{@{}clrrrrrrr|rrr@{}}
\toprule
\multicolumn{1}{l}{\multirow{2}{*}{}} & \multirow{2}{*}{} & \multicolumn{1}{c}{\multirow{2}{*}{\aspectnewstiny{}}} & \multicolumn{1}{c}{\multirow{2}{*}{\debatesumtiny{}}} & \multicolumn{1}{c}{\multirow{2}{*}{\scifacttiny{}}} & \multicolumn{1}{c}{\multirow{2}{*}{\openasptiny{}}} & \multicolumn{1}{c}{\multirow{2}{*}{\saliencytiny{}}} & \multicolumn{1}{c}{\multirow{2}{*}{\evidencedetectiontiny{}}} & \multicolumn{1}{c|}{\multirow{2}{*}{\textbf{Avg}}} & \multicolumn{3}{c}{\textbf{Token-level}} \\
\multicolumn{1}{l}{} &  & \multicolumn{1}{c}{} & \multicolumn{1}{c}{} & \multicolumn{1}{c}{} & \multicolumn{1}{c}{} & \multicolumn{1}{c}{} & \multicolumn{1}{c}{} & \multicolumn{1}{c|}{} & \multicolumn{1}{c}{P} & \multicolumn{1}{c}{R} & \multicolumn{1}{r}{$F_1 \pm$ CI} \\ \midrule
\multirow{5}{*}{\textbf{\makecell[cc]{\textbf{\rotatebox[origin=c]{90}{transfer-}} \textbf{\rotatebox[origin=c]{90}{only}}}}} & \llamaIcl{} & 34.6 & \textbf{46.9} & 44.8 & 28.6 & \textbf{42.9} & 13.5 & 35.2 & 42.3 & 51.0 & 41.2 $\pm$ 1.5 \\
 & \llama{} & 29.4 & 42.4 & 47.5 & 41.9 & 36.6 & 27.3 & 37.4 & 44.0 & 50.5 & 41.9 $\pm$ 1.6 \\  \cdashline{2-12}  
 & + \loo{} & 34.0 & 29.1 & 25.8 & 30.4 & 41.9 & 10.2 & 28.6 & 24.0 & \textbf{61.9} & 29.8 $\pm$ 1.5 \\
 & + \igcsriunion{} & {\ul \textbf{37.0}} & 36.7 & 42.6 & {\ul \textbf{49.3}} & 37.5 & {\ul 33.6} & \textbf{39.5} & 45.7 & 56.4 & \textbf{45.7} $\pm$ 1.7 \\
 & + \igcsrimajority{} & 35.6 & 25.2 & \textbf{48.1} & 47.1 & 32.4 & {\ul \textbf{35.6}} & 37.3 & \textbf{50.0} & 46.3 & 43.2 $\pm$ 1.7 \\ \midrule
\multirow{4}{*}{\textbf{\makecell[cl]{\textbf{\rotatebox[origin=c]{90}{\footnotesize{supervision+}}} \textbf{\rotatebox[origin=c]{90}{\footnotesize{transfer}}}}}} & \llamaSup{} & 40.6 & 63.5 & 66.0 & -- & -- & -- & -- & \multicolumn{1}{l}{--} & \multicolumn{1}{l}{--} & \multicolumn{1}{l}{--} \\ \cdashline{2-12} 
 & + \loo{} & {\ul 42.3} & \textbf{64.1} & {\ul 70.0} & -- & -- & -- & -- & \multicolumn{1}{l}{--} & \multicolumn{1}{l}{--} & \multicolumn{1}{l}{--} \\
 & + \igcsriunion{} & {\ul 42.7} & 63.7 & \textbf{{\ul 72.1}} & -- & -- & -- & -- & \multicolumn{1}{l}{--} & \multicolumn{1}{l}{--} & \multicolumn{1}{l}{--} \\
 & + \igcsrimajority{} & \textbf{{\ul 43.2}} & 63.6 & 68.8 & -- & -- & -- & -- & \multicolumn{1}{l}{--} & \multicolumn{1}{l}{--} & \multicolumn{1}{l}{--} \\ \specialrule{0.3pt}{0pt}{0.4pt} \specialrule{0.3pt}{0pt}{\belowrulesep}
\multirow{5}{*}{\textbf{\makecell[cl]{\textbf{\rotatebox[origin=c]{90}{prompt-based}} \textbf{\rotatebox[origin=c]{90}{models}}}}} & \claude{} & 31.3 & 49.6 & 54.5 & 52.2 & 43.7 & 28.2 & 43.2 & 49.0 & 58.4 & 47.4 $\pm$ 1.7 \\
 & \llamaM{} & 29.3 & 40.7 & 58.5 & 56.8 & 33.2 & 44.9 & 43.9 & 55.7 & 49.4 & 47.8 $\pm$ 1.7 \\
 & \llamaL{} & 30.0 & 45.4 & 56.2 & 59.8 & 35.1 & 42.1 & 44.7 & 51.6 & 57.1 & 49.3 $\pm$ 1.8 \\
 & \gpt4{} & 32.8 & 39.0 & 58.6 & 57.4 & 39.1 & 50.1 & 46.2 & 60.0 & 53.9 & 51.4 $\pm$ 1.9 \\
 & \gptIcl{} & 33.9 & 45.5 & 57.3 & 55.0 & 39.9 & 47.5 & 46.5 & 59.7 & 55.5 & 52.2 $\pm$ 1.7 \\ \bottomrule
\end{tabular}%
}
\caption{
Performance on \igcsbench{} tasks (\S\ref{subsec:results-main}) using their original evaluation metrics (left) and overall \textbf{Token-level} metrics (right), comparing \textit{transfer-only} (top), \textit{supervision+transfer} (middle), to prompt-based methods (bottom).
In the two topmost sections, the baseline(s) appear above the dashed line, and the 
transfer configurations of fine-tuned Llama-3-8B variants (\S\ref{subsec:ft-models}) below it.
In the \textit{transfer-only} section, only \ri{} is used for fine-tuning, whereas in the \textit{supervision+transfer} section, the task-specific training set is also included in the fine-tuning mix, when the task has such a set.
For each task in each section, \textbf{bold} indicates the highest score, and \uline{underline} indicates statistical significance ($p < 0.05$) with respect to the baselines in each section.
The four rightmost columns report the overall average score across the six tasks (\textbf{Avg}) and token-level \textbf{P}recision, \textbf{R}ecall, and $F_1$ $ \pm $ confidence interval ($\alpha=0.05$).
}
\label{tab:main-results}
\end{table*}

\subsection{Inference-time Configurations}
\label{subsec:models-infer-config}
When employing LLMs for selecting text spans from their input, two issues arise. First, for multi-text inputs, the input context length, as well as the output length, may become challenging \cite{liu-etal-2024-lost,levy-etal-2024-task}. Second, as mentioned in \S\ref{subsec:background-cs-models}, while instructed to copy verbatim the selected source spans, LLMs sometimes deviate from the source, e.g. omitting words, generating paraphrases, or hallucinating. We next address these two issues.

\paragraph{Document-level inference.}\label{subsec:model-frag}
Given multiple documents as input, the typical approach would be to concatenate all of them in a single prompt. However, we observed that when both the input and the expected output are relatively long, models tend to produce shorter outputs than required, decreasing performance (\S\ref{subsec:results-inference-config}). Yet, in a content selection task, the output for a multi-document input 
could simplistically be viewed as
the concatenation of selections from all documents. Hence, we experimented with prompting the model separately with each input document, then concatenating all output selections. While in this approach selection decisions for each document cannot consider information from other documents, we found it to perform better overall thanks to the shorter inputs and outputs, and hence adopted it in our modeling.

\paragraph{Grounding the output to the source text.}
\label{subsec:model-grounding}

Addressing the second issue above, where the output selections generated by the model might deviate from the source text, we ground the model output back to the source. In case an exact match is not found, we exhaustively search for the closest source span in terms of token-level Levenshtein distance\repl{, allowing for a a maximal distance of 3}{} (technical details in \autoref{app:models-grounding}).


\section{Results and Analysis}\label{sec:experiments}

Our results and analysis assess the utility of transfer learning based on our unified content selection scheme (\S\ref{subsec:results-main}), the effectiveness of document level inference (\S\ref{subsec:results-inference-config}), and the generality of our proposed generic evaluation metric (\S\ref{subsection:results-meta-eval}).

\subsection{Transfer-learning of Fine-tuned models}\label{subsec:results-main}

\autoref{tab:main-results} presents the results of applying our various transfer learning configurations, including \textit{transfer-only} and \textit{supervised+transfer} (\S\ref{subsec:ft-models}), as well as results for larger prompt-based models (\S\ref{subsection:prompt-models}), tested over the six \igcsbench{} datasets.
The results of each task are measured with the corresponding evaluation metric of the original task dataset, while overall scores, averaged over all datasets (as explained in \S\ref{subsec:overall-scores}), are presented in the right-hand side of the table.
\autoref{tab:app-main-results-detailed} in App. \ref{app:detailed-tl-results} analogously presents the scores measured by token-level $F_1$, which we advocate as a generic content selection measure, yielding very similar trends.

\subsubsection{\textit{Transfer-only} Configurations}
\label{par:results-main-zeroshot}

The top section of \autoref{tab:main-results} presents the results obtained in the absence of training data for the targeted tested task. 
The first two rows provide the baselines in this setting, namely zero-shot and few-shot prompting, without any fine-tuning on transfer data. 
The 3rd row corresponds to leave-one-out fine-tuning over the \igcsbench{} training data 
(i.e., testing on a task when the model is fine-tuned with the \textit{other} tasks' train sets). Rows 4-5 correspond to fine-tuning with \igcsri{}.

As can be seen in the +\loo{} row, performance degrades substantially, relative to the prompt-based baselines, when transferring training data from a few specific content selection tasks to another task. In such a setting, the model seems to be steered toward distinct use-cases, which fails to generalize to different tasks.
On the other hand, when fine-tuning with our \textit{generic} \igcsri{} dataset, the fine-tuned models outperform the prompt-based baselines with an overall $F_1$ score of 45.7 compared to the highest scoring baseline with 41.8.

Additionally, the overall recall score of \igcsriunion{} and precision score of \igcsrimajority{} are at least 5.4 and 6.0 points higher than both baselines, respectively.
This outcome is expected, as the union merging strategy encourages the fine-tuned model to select more tokens than the majority strategy, which is more conservative and therefore induces higher precision.
Thus, the two \igcsri{} fine-tuned variants offer complementary trade-offs, allowing one to be chosen when a given task prioritizes precision over recall or vice versa.
For example, as shown in the task-specific results (left part of the table), \debatesumshort{}, for which selection size is relatively long, benefits more from \igcsriunion{} than from \igcsrimajority{}. On the other hand,  \igcsrimajority{} provides greater benefit for \scifactshort{}, for which the expected selection size if relatively short (favoring precision).
Overall across the benchmark, the union variant is more advantageous than the majority variant.
Further, when examining the per-task results, we find that two out of the six tasks do not benefit from transfer-only finetuning. This  suggests that while our results indicate that such fine-tuning is beneficial overall, when developing a generic content selection model, the utility of such fine-tuning should be verified when targeting a specific task.
As a reference point, \autoref{tab:bench-tasks-desiderata} in \autoref{app:bench-details} compares the results of transfer learning from our \ri{} datasets to the analogous previously reported result for each dataset.

To more broadly investigate the advantages of fine-tuning with \ri{} compared to prompt-based methods, we report overall $F_1$ scores for seven additional models in \autoref{fig:bars_overall_scores}. Across all models evaluated, the fine-tuned variants consistently outperform their prompt-based counterparts. While, as may be expected, the smallest models exhibit the greatest gains from \igcsri{} (transfer) finetuning, the larger models we tested also exhibit significant gains of several $F_1$ points.

To further assess the reliability of these results, we repeated each model run two more times, with two additional prompt variants that were derived automatically from the original human-written instruction, following a common-practice LLM-based prompt tuning, as elaborated in \autoref{app:prompt-robustness}. The results across these runs, presented in \autoref{fig:prompt_robustness} in the appendix, consistently show similar trends exhibited in \autoref{fig:bars_overall_scores}, while exhibiting some variations in absolute performances, demonstrating the typical sensitivity of LLMs to prompt variants \cite{DBLP:conf/iclr/Sclar0TS24, mizrahi-etal-2024-state}.

To summarize, the results thus far suggest that fine-tuning with a generic dataset of diverse content selection scenarios allows the model to better generalize to different tasks, showing the potential value of our synthetic dataset when addressing tasks for which targeted training data is absent.

\begin{figure*}[t]
    \centering
    
    \includegraphics[width=1.0\textwidth]{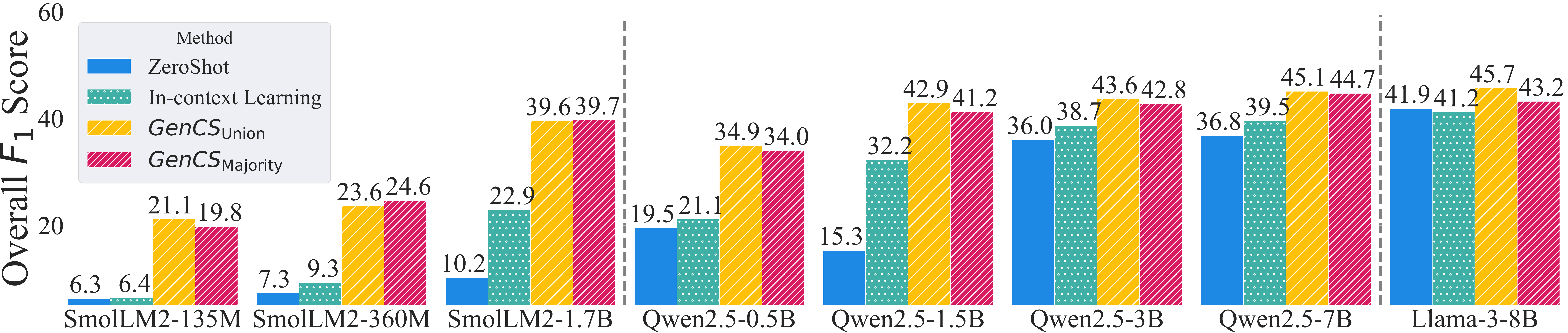}
    \caption{Overall $F_1$ scores on \igcsbench{} for four methods across eight small language models of up-to 8B parameters from Qwen-2.5, SmolLM2 and Llama-3 families. All tested models benefit from fine-tuning with \ri{}, especially smaller ones. 
    The largest confidence interval across models is 2.0 ($\alpha = 0.05$).}
    \label{fig:bars_overall_scores}
\end{figure*}

\subsubsection{\textit{Supervision+transfer} Configurations}
\label{par:results-main-supervised}
The middle section of \autoref{tab:main-results} presents results for the setting where training data \textit{is} available for the tested task. The first row in this section provides the baseline for this setting, namely the zero-shot model (2nd row in the top section) fine-tuned with the training data of the tested task (available for three tasks). Naturally, such task-specific fine-tuning
improves results for all three tasks (comparing the baseline rows of the two sections). 

Notably, results improve by enriching the fine-tuning data with transfer training sets (second row of the section), and statistically significantly so (see \autoref{app:significance-testing} for details) for two of the three tasks in this setting. The magnitude of improvement is somewhat smaller than in the transfer-only setting, 
which is seemingly anticipated, since the transfer-training data supplements the existing task-specific training data.
Interestingly, we observe that while transfer learning from different tasks was not helpful in the transfer-only setting (\S\ref{par:results-main-zeroshot}), it does improve performance for all three tasks when it is combined with the available task-specific training data.
Finally, our \igcsri{} datasets prove beneficial in this setting as well.

\subsubsection{Larger Prompt-based Models}
\label{par:results-main-prompt}
The bottom section of \autoref{tab:main-results} provides the results for larger models using prompt-based approaches, as reference points.
These models mostly outperform the smaller 8B models in the transfer-only version, as can be expected when no task-specific training data is available for the targeted task. Still, transfer learning offers value when access to large models is limited or cost-prohibitive.
When task-specific training data \textit{is} available for fine-tuning, smaller models notably outperform larger models, and this performance gap widens further when transfer learning using \ri{} is applied.

Taken together, our findings suggest that the proposed generic IGCS scheme, combined with transfer learning on our datasets, offers an effective and appealing approach for a variety of content selection tasks and use cases.

\subsection{Ablation Analysis}


\paragraph{Synthetic dataset generation configurations.}\label{subsec:gencs-ablation-results}

To test the impact of various design choices in our automatic dataset generation process (see \S\ref{sec:ri-dataset}), we generated several ablated variants of this process and tested the performance of 3 models when finetuned with the different dataset variants.
Specifically, we created four variants of the \ri{} dataset (details in \autoref{app:synthetic-pipeline-ablation}): 
(1) \riSingleStep{} --- combining steps 1 and 2 such that an instruction and its selection are generated with a single prompt;
(2) \riSingleInst{} --- generating only a single instruction in step 1;
(3) \riSingleModel{} --- using only a single model in step 2;
and (4) \igcsriunion{} --- the full process used to generate the \igcsriunion{} dataset, as described in \S\ref{sec:ri-dataset}.
Then, we fine-tuned three models on each dataset variant.
\autoref{tab:gencs-pipeline-ablation} presents the results of each finetuning variant, in comparison to the best prompt-based (zero-shot or in-context) baseline for the respective model (row~1). As shown, all fine-tuned models achieved higher overall performance on \igcsbench{} than the baseline, where two models performing best with the full configuration.

\paragraph{Document-level inference.}\label{subsec:results-inference-config}

We next analyze the impact of document-level inference (\S\ref{subsec:models-infer-config}) over the four \igcsbench{} tasks that have multi-document inputs, shown in \autoref{fig:bar_fragment_diff}. 
To that end, we measure model performance on these tasks when feeding the model the full input versus feeding it document by document and then concatenating all document-level selections (\S\ref{subsec:model-frag}). 
For each task, we measured the average performance of 9 models (details in App. \ref{app:detailed-frag-results}) on the task for each of the two settings (single vs. multi-document input), and plotted the difference between these averages in \autoref{fig:bar_fragment_diff}. 
Notably, performance for \openaspshort{} and \saliencyshort{} improves substantially, across all 9 models. Meanwhile, performance is negligibly affected for \scifactshort{}, and slightly degrades for \evidencedetectionshort{}. 
The differentiating factor seems to be the output \textit{selection size}, where the model struggles to generate sufficient selections when processing all input documents at once. Processing one document at a time, on the other hand, triggers the model to generate more complete selections for each document, and hence for all documents as a whole.
Further analysis appears in App. \ref{app:detailed-frag-results}.

\begin{table}[t]
\resizebox{\columnwidth}{!}{%
\begin{tabular}{@{}lrrr@{}}
\toprule
  & \multicolumn{1}{c}{SmolLM2-1.7B} & \multicolumn{1}{c}{Qwen2.5-7B} & \multicolumn{1}{c}{Llama-3-8B} \\ \midrule
Prompt-based & 22.9 $\pm$ 1.4 & 39.5 $\pm$ 2.0 & 41.9 $\pm$ 1.6 \\ \hdashline
\riSingleStep{} & 32.9 $\pm$ 1.5 & 47.0 $\pm$ 1.6 & 44.7 $\pm$ 1.6 \\
\riSingleInst{} & 34.2 $\pm$ 1.4 & 46.5 $\pm$ 1.7 & 44.1 $\pm$ 1.8 \\
\riSingleModel{} & 38.2 $\pm$ 1.5 & 44.7 $\pm$ 1.6 & 43.1 $\pm$ 1.8 \\
\igcsriunion{} & 39.6 $\pm$ 1.6 & 45.1 $\pm$ 1.7 & 45.7 $\pm$ 1.7 \\\bottomrule
\end{tabular}%
}
\caption{
Overall $F_1$ scores $ \pm $ confidence intervals ($\alpha=0.05$) on \igcsbench{} of three fine-tuned models with training sets obtained using different synthetic pipeline configurations (\S\ref{sec:ri-dataset}). All fine-tuned models outperform the best prompt-based setting (first row), while different pipeline configurations are more effective for different models.}
\label{tab:gencs-pipeline-ablation}
\end{table}
\begin{figure}[t]
    \centering
    \includegraphics[width=\columnwidth]{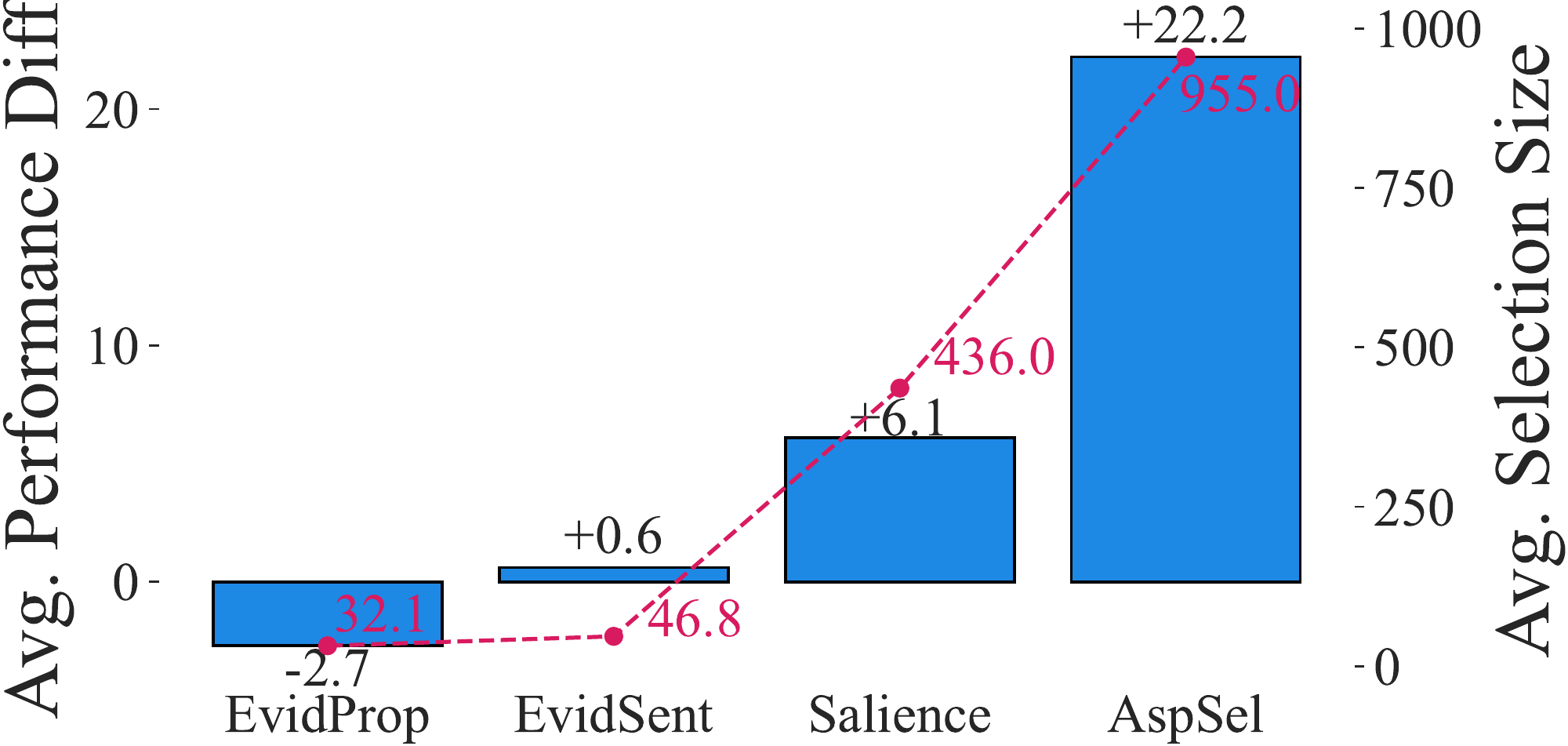}
    \caption{Difference in original metric scores between applying document-level inference (\S\ref{subsec:model-frag}) and processing document-sets as a whole, averaged across 9 models.
    The red line shows the average \textit{selection size} (\S\ref{subsec:cs-properties}) per task. It seems that document-level inference is more beneficial when a task has a higher selection size.}
    \label{fig:bar_fragment_diff}
\end{figure}

\subsection{Assessing a Generic Evaluation Metric}
\label{subsection:results-meta-eval}

Next, we assess our proposal to use token-level $F_1$ as a generic evaluation metric for content selection tasks (\S\ref{subsection:evaluation}).
As mentioned earlier, each of the six \igcsbench{} tasks has been originally evaluated using its own metric: while \saliencyshort{} and \evidencedetectionshort{} already employed token-level $F_1$, the other four tasks employed different metrics
(see App. \ref{app:bench-dataset-details} for details). 
To assess the generality of token-level $F_1$, we measured its system level correlation
with the other four metrics (see \autoref{tab:meta-eval-results} in App. \ref{app:meta-eval}).
We find that the token-level $F_1$ measure exhibits strong or very strong correlation with the all other metrics, suggesting its suitability as a generic evaluation metric for content selection tasks.

\paragraph{Overall score.}

To evaluate the effectiveness of the overall score based on token-level $F_1$, we computed its correlation with the overall score derived from the original task-specific metrics (described in \S{}\ref{subsection:evaluation}). The correlation considers the scores of all models and methods presented in \autoref{fig:bars_overall_scores}. We find that the two variants exhibit an almost perfect correlation, with Pearson's $r > 0.99$. This suggests that the overall score based on the generic token-level metric is a reliable indicator of model performance on the \igcsbench{} benchmark.

\section{Conclusion}\label{sec-conclusions}

We introduced \textit{instruction-guided content selection (IGCS)}, a unified scheme that generalizes a broad range of content selection tasks by encoding the task objective and input as a natural language instruction. To support this framework, we developed the first unified benchmark for this setting, \igcsbench{}, and an extensive generic synthetic dataset for training, \igcsri{}, which covers diverse content selection requests. 
Notably, we showed that leveraging these datasets for transfer learning is often effective, whether training data for the specific targeted task is available or not.
Additionally, we propose document-level inference to circumvent the shortcomings of large language models when addressing content selection for long contexts.
Finally, we proposed using token-level $F_1$ evaluation as a standard generic metric for content selection, showing that it strongly correlates with prior task-specific metrics.
Overall, we suggest the utility of our framework for future modeling of diverse content selection tasks, while paving the way for future research to model ad-hoc user-generated content selection instructions.

\section{Limitations}\label{sec-limitations}

\igcsbench{} is built upon six particular content selection tasks. While these tasks are shown to be diverse, an alternative set of tasks may behave differently in terms of transfer learning, and lead to slightly different findings.

Throughout our study we experimented with several prompt variants, yet it is still possible that better prompts exist.

Finally, there is a lack of detailed documentation regarding the pre-training data used by the tested models. This makes it challenging to determine whether our test data is included in their training corpora, raising the possibility of data contamination.

\section*{Acknowledgments}
We thank the reviewers and the Action Editor for their constructive feedback, which substantially improved this manuscript.
This work was supported by Israel Science Foundation grant 2827/21.


\bibliography{custom}
\bibliographystyle{acl_natbib}

\appendix\label{sec:appendix}











\section{\igcsbench{} Details}
\label{app:bench-details}\label{app:bench-dataset-details}

In this section we provide details on creating \igcsbench{}, introduced in Section \ref{sec:bench-dataset}, based on the six CS datasets.
We split every dataset into train, development and test sets except for our two datasets from \citet{ernst2024power} that only include a test set.
The full prompt template that we drafted for the tasks is shown in \autoref{fig:app_prompt_igcs}.

\autoref{tab:bench-tasks-desiderata} presents the desiderata for the tasks. Specifically, it displays: (1) model scores for each task, as reported in the respective paper. We find that there is substantial \textbf{room for improvement} on these datasets. (2) Inter annotator agreement of the annotations when the datasets were curated. The high Cohen's $\kappa$ scores are an indication for \textbf{the quality of the data}.

\paragraph{Evidence Retrieval (\scifactshort{}).} We randomly split the original train set from \scifact{} \cite{wadden_fact_2020} into train and a development sets with 687 and 122 instances, respectively.
The original development set is used as our test set, since the original \scifact{} test set is gated behind a leaderboard submission system.

For evaluation, we followed the original paper and used sentence-level $F_1$. Since the model selects tokens, we first identify the sentences that contain them, which are then used for computing the sentence-level metric.

\paragraph{Salience and Proposition-level Evidence Detection (\saliencyshort{} and \evidencedetectionshort{}).} From the \spark{} \cite{ernst2024power} dataset, we only utilize the respective test sets. \citet{ernst2024power} originally used an automatically derived and lower quality training set.
\saliencyshort{} and \evidencedetectionshort{} include 98 and 1,332 instances in their test sets, respectively.

For evaluation, we followed the original paper and used token-level $F_1$ for both tasks. This is the same as the generic metric that we propose to apply, in \S\ref{subsection:evaluation}.

\paragraph{Aspect-based Sentence Selection (\openaspshort{}).}
\openasp{} \cite{amar-etal-2023-openasp} defines a sentence selection task.
We use the original test set with 27 samples and split the original development set into training and development sets with 13 and 11 samples, respectively.

For evaluation, we follow the originally proposed sentence-level micro-$F_1$ metric.
Similar to the case of \scifactshort{}, we first identify the sentences of the selected tokens, and then compute the measurement.

\paragraph{Extractive Aspect-based Summarization (\aspectnewsshort{}).}
For the extractive aspect-based summarization task (single document), \citet{ahuja_aspectnews_2022} annotated 100 documents from two topics, each with two aspects, yielding a total of 400 document-aspect-summary instances.
We use the \textit{Fraud} topic and its two aspects -- \textit{penalty} and \textit{nature} -- as the test set, and split the remaining 100 documents from the Earthquake topic into 160 training and 40 development instances.
\aspectnews{} has 5 annotations per document; we retain these as 5 separate gold references and evaluate them following the multiple-reference selection approach (\S\ref{subsection:evaluation}).

For evaluation, we use
the metric from the
original paper, which computes sentence-level $F_1$ against references with soft labels.
As in the other sentence-level evaluations, the sentences used for the evaluation are those that contain the tokens selected by a model.

\paragraph{Argument Mining (\debatesumshort{}).} \debatesum{} \cite{roush_debatesum_2020} reported results on a test set of 18,738 instances but did not provide the original dataset splits, as confirmed in both our review and the project's official repository.\footnote{\url{https://github.com/Hellisotherpeople/DebateSum/issues/3}} To support reproducible research, we propose new training, development, and test splits for \debatesum{}. The dataset, originally sourced from the debate.cards website, spans 2013--2019 with a total of 187,386 instances. We allocate years 2013--2016 for training, 2017 for development, and 2018--2019 for testing. To avoid contamination, we filter instances by removing any that share identical abstract summaries, extractive summaries, or full document fields across splits. For \igcsbench{}, we randomly sample 1,000 instances from each split.

For evaluation, we followed the original paper and used ROUGE-2-$F_1$ as our primary metric.

\section{Computing Confidence Intervals for Token-level $F_1$}\label{app:bootstrap-overall-score}

We estimated 95\% confidence intervals via bootstrap resampling \cite{efron1994bootstrap} with 10,000 iterations. At each iteration, we drew instances with replacement from each of the six \igcsbench{} datasets and computed the corresponding $F_1$ score for that bootstrap sample.

\begin{figure}[tb]
    \begin{tcolorbox}[colback=black!5!white, colframe=black, coltitle=white, title=\igcsbench{} Prompt Template, fonttitle=\bfseries,
      boxsep=1pt,
      left=1pt,
      right=1pt,
      top=1pt,
      bottom=1pt,
      before upper={\vspace{1pt}},
      after upper={\vspace{1pt}},
      parskip=0pt
    ]
        \textbf{\{\circlet{1} instruction\}}. Output the exact \textbf{\{\circlet{2} selection\_type\}} from the given \textbf{\{\circlet{3} source\_type\}} as a valid json array of strings. Do not change the copied text. \\
    
    document \#0: \\ Document 0 text\dots \\
    document \#1: \\ Document 1 text\dots \\
    \dots
    \end{tcolorbox}
    \caption{The prompt template for \igcsbench{} tasks.
    The \texttt{\circlet{1} instruction} for each task is detailed in \autoref{tab:igcs-instructions}. The \texttt{\circlet{2} selection\_type} is \textit{``sentences"} for \openaspshort{}, \aspectnewsshort{}, \scifactshort{}, and \textit{``text phrases"} for the other tasks.
    The \texttt{\circlet{3} source\_type} varies by task: \textit{``document"} for single-document tasks (\aspectnewsshort{} and \debatesumshort{}), \textit{``abstract(s)"} for \scifactshort{}, and \textit{``documents"} for the remaining three multi-document tasks.}
    \label{fig:app_prompt_igcs}
\end{figure}
\begin{table}[t]
\centering
\resizebox{\columnwidth}{!}{%
\begin{tabular}{@{}lrrr@{}}
\toprule
Task                 & \multicolumn{1}{c}{Cohen's~$\kappa$}      & \multicolumn{1}{c}{Best-reported}  & \multicolumn{1}{c}{Best-\ri{}} \\  \midrule
\aspectnewsshort{}                                                & \textsuperscript{\textdagger{}}-  & 45.0     & 37.0                           \\
\debatesumshort{}                                                 & -     & \textsuperscript{\textdaggerdbl{}}38.5      & 36.7      \\
\scifactshort{}                                                   & 0.71  & 44.0        & 48.1                        \\
\openaspshort{}                                                  & 0.64   & 34.4     & 49.3                          \\
\saliencyshort{}                                                  & 0.72  & 31.0        & 37.5                        \\
\evidencedetectionshort{}                                         & 0.72  & 32.0        & 35.6                        \\ \bottomrule
\end{tabular}%
}
\caption{
Reported inter-annotator agreement and performance figures for the original datasets incorporated in our \igcsbench{} benchmark (whose details appear in \S\ref{subsection:bench_tasks}). The Cohen's~$\kappa$ figures specify the reported inter-annotator agreement levels for the respective dataset, when available. 
Best-reported are the previously reported results for each dataset, 
using the original evaluation measure proposed for each dataset. We quote here  performance figures in the transfer learning setting (except for \aspectnewsshort{}, see dagger), as this is the setting on which we focus in this paper, with respect to the utility of our generic \ri{} training dataset.
Finally, for comparison, Best-\ri{} presents the best performance obtained in our experiments when utilizing our \ri{} training set in the transfer-only setting (maximum value among rows 4 and 5 in \autoref{tab:main-results}).
As shown in the table, our transfer-only results improve over the prior performances for 4 of the datasets (while direct comparison is not available for \debatesumshort{}, see double-dagger). 
\\
\textdagger{}: 
Inter-annotator agreement was not reported for this task due to its subjectivity; instead, the dataset includes five reference selections per instance, where model performances are computed against each of them.\\
\textdaggerdbl{}: Here we quote the best supervised results, rather than transfer learning results, since the latter setting was not previously attempted for this dataset. Further, the original test split for this dataset  was not released, hence this figure is not fully comparable to our result (in the last column), which was computed on our introduced test split.
}
\label{tab:bench-tasks-desiderata}
\end{table}

\section{\igcsri{} Details}\label{app:ri-dataset-details}

\subsection{Data Collection}

From Multi-News \cite{fabbri-etal-2019-multi}, English Wikipedia, PubMed, books \cite{DBLP:conf/iclr/RaePJHL20}, and hotel reviews \cite{10.1145/1835804.1835903}, we uniformly sampled 500 documents, each with between 350 and 3500 tokens, as counted with \texttt{nltk.word\_tokenize}.
From Enron \cite{10.1007/978-3-540-30115-8_22}, we sampled 250 email threads containing multiple emails and 250 threads with a single email.

From GitHub, we sampled one million source code files in one of the following 15 programming languages: \texttt{Assembly, C, C\#, C++, GO, Java, JavaScript, PHP, Perl, Python, Ruby, Rust, Scala, Shell, TypeScript}, each with a permissive license of either \texttt{APACHE\-2.0} or \texttt{MIT}.
Next, we grouped files from the same repository and folder into multi-source task instances.
Finally, we sampled 250 multi-source task instances and 250 single-instance source files, resulting in 500 GitHub samples.

\subsection{Synthesizing Instructions and Annotating Selections}\label{app:ri-dataset-prompts}

For synthesizing instructions (Step 1 in \S\ref{subsec:ri-dataset-generation}), the prompt is presented in \autoref{fig:app_prompt_ri_annotation_instructions}. For synthesizing possible content selections (Step 2), the prompt is presented in \autoref{fig:app_prompt_ri_annotation_selections}.

We intentionally under-specified the definition of an instruction to encourage the generation of diverse instructions.

\section{Synthetic Pipeline Configuration Variants Details}\label{app:synthetic-pipeline-ablation}

In \autoref{tab:gencs-pipeline-ablation}, we compare models fine-tuned on different pipeline configurations. For \riSingleStep{}, we used \gpt4{} as the annotator, issuing a single prompt of combined instruction and guidelines from \autoref{fig:app_prompt_ri_annotation_instructions} and \autoref{fig:app_prompt_ri_annotation_selections}. 
We utilized the annotations for \ri{} to derive \riSingleModel{} and \riSingleInst{}.
For \riSingleModel{}, we used \gpt4{} as the single selection model, while for \riSingleInst{}, we used only the first generated instruction.

\begin{figure}[tb]
    \begin{tcolorbox}[colback=black!5!white, colframe=black, coltitle=white, title=\igcsri{} Instruction Generation Prompt, fonttitle=\bfseries,
      fontupper=\scriptsize,
      fontlower=\scriptsize,
      boxsep=1pt,
      left=1pt,
      right=1pt,
      top=1pt,
      bottom=1pt,
      before upper={\vspace{1pt}},
      after upper={\vspace{1pt}},
      parskip=0pt
    ]
        \textbf{System:} You are a manager of a \texttt{\{software|publishing\}} firm \\

        \textbf{User:} You are a manager of a \texttt{\{software|publishing\}} firm and you are required to train the best students on how to perform \texttt{\{code|content\}} selection from given sources. \\
Write 5 short instructions for selecting \texttt{\{code|content\}} from the given \texttt{\{file|document\}}(s) to challenge students and train them on how to select relevant \texttt{\{code|content\}} based on diverse instructions. \\

Guidelines:
\begin{enumerate}[topsep=0pt,itemsep=-1ex,partopsep=1ex,parsep=1ex]
  \item You must keep the 5 instructions short and concise - as a single sentence.
  \item Instructions must start with the words "Select \texttt{\{code|content\}}" as they are always for selecting \texttt{\{code|content\}} from the \texttt{\{file|document\}}(s) and never for writing a new \texttt{\{code|text\}} nor paraphrasing the original content.
  \item Instructions should not be too specific that hint on the answer and not too vague that cannot be fulfilled.
  \item Write the instructions as a numbered list.
\end{enumerate}

    \texttt{\{Source File|Document\}} \#0: \\ Document 0 text\dots \\
    \texttt{\{Source File|Document\}} \#1: \\ Document 1 text\dots \\
    \dots
    \end{tcolorbox}
    \caption{The prompt template for annotating IGCS instructions is as follows: from each choice of \texttt{\{\circlet{1}|\circlet{2}\}}, we use \circlet{1} for the GitHub code dataset and \circlet{2} for all other datasets.
    For empty selection annotation, the second sentence is changed as follows: ``\textit{Write 5 short instructions that have no matching \texttt{\{code|content\}} from the given \texttt{\{file|document\}}(s) to challenge students and train them to avoid selecting \texttt{\{code|content\}} when none matches the instruction.}".
    }    \label{fig:app_prompt_ri_annotation_instructions}
\end{figure}
\begin{figure}[tb]
    \begin{tcolorbox}[colback=black!5!white, colframe=black, coltitle=white, title=\igcsri{} Selection Annotation Prompt, fonttitle=\bfseries,
      fontupper=\scriptsize,
      fontlower=\scriptsize,
      boxsep=1pt,
      left=1pt,
      right=1pt,
      top=1pt,
      bottom=1pt,
      before upper={\vspace{1pt}},
      after upper={\vspace{1pt}},
      parskip=0pt
    ]
        \textbf{System:} You are a helpful assistant. \\

        \textbf{User:} For every instruction listed below, select \texttt{\{code|content\}} from the below \texttt{\{source file|document\}}(s) that matches the instruction.

Guidelines:
\begin{enumerate}[topsep=0pt,itemsep=-1ex,partopsep=1ex,parsep=1ex]
  \item Output the exact verbatim \texttt{\{code|text phrases\}} from the \texttt{\{source file|document\}}(s). Do not change spaces or punctuation, do not fix typos and avoid any other changes to the \texttt{\{code|content\}} you select.
  \item Follow the instructions as closely as possible and pay careful attention to them.
  \item Output format must be a two level nested list, the first level is the instruction and the second level is the multiple \texttt{\{code|content\}} selections copied from the original \texttt{\{source file|document\}}(s).
\end{enumerate}

Instructions:
    \begin{enumerate}[topsep=0pt,itemsep=-1ex,partopsep=1ex,parsep=1ex]
        \item \textit{<instruction \#1 text>}.
        \item \textit{<instruction \#2 text>}.
        \item \textit{<instruction \#3 text>}.
        \item \textit{<instruction \#4 text>}.
        \item \textit{<instruction \#5 text>}.
    \end{enumerate}

    \texttt{\{Source File|Document\}} \#0: \\ Document 0 text\dots \\
    \texttt{\{Source File|Document\}} \#1: \\ Document 1 text\dots \\
    \dots
    \end{tcolorbox}
    \caption{The prompt template for annotating selections. From every choice of \{\circlet{1}|\circlet{2}\}, we use \circlet{1} for the GitHub code dataset and \circlet{2} otherwise.
    }
    \label{fig:app_prompt_ri_annotation_selections}
\end{figure}

\section{\igcsri{} Manual Quality Assessment}
\label{app:ri-qa}

The guidelines for rating the intructions' \textit{naturalness} and \textit{specificity} are in \autoref{fig:annotation-guidelines}.
The annotation interface for selecting spans based on instructions is presented in \autoref{fig:qa_annotation_interface}.

\begin{figure}[tb]
    \centering
    \includegraphics[width=1.0\columnwidth]{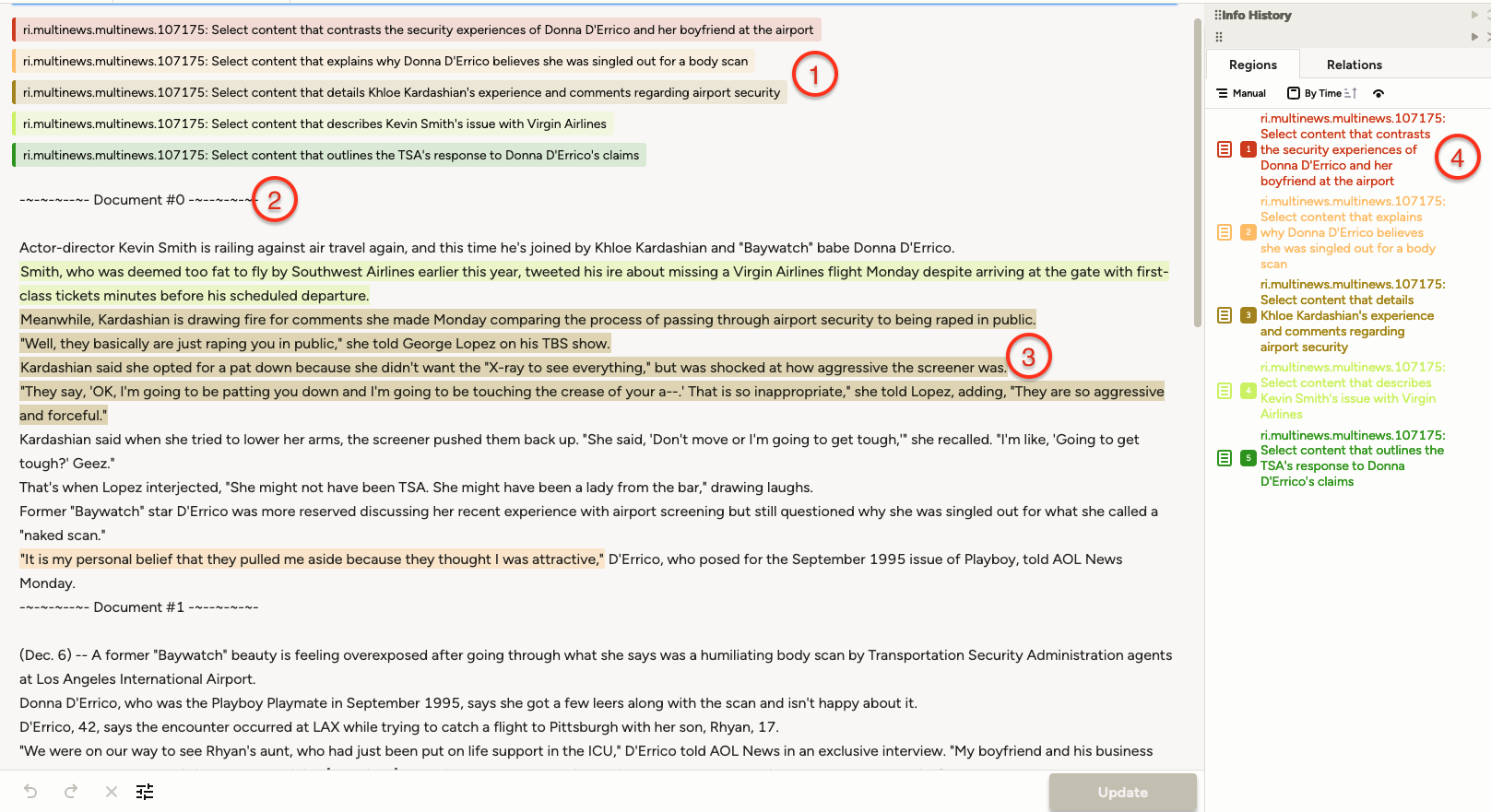}
    \caption{Our annotation interface for manual content selection, based on Label Studio (\url{https://labelstud.io}).  In \circlet{1}, the five instructions for the document set can be selected in order to enable highlighting of text spans in the source \circlet{2}, such as the highlighted span shown in \circlet{3}. The highlights are added to the selected spans panel marked in \circlet{4}. The text within the figure is for illustrative purposes only and does not need to be read.
    }
    \label{fig:qa_annotation_interface}
\end{figure}

\begin{figure*}[tb]
    \centering
    \begin{tcolorbox}[colback=black!5!white, colframe=black, coltitle=white, title=\igcsri{} Manual Annotation Guidelines for Instructions and Selections, 
        fonttitle=\scriptsize\bfseries,
  fontupper=\scriptsize,
  fontlower=\scriptsize,
  boxsep=1pt,
  left=1pt,
  right=1pt,
  top=1pt,
  bottom=1pt,
  before upper={\vspace{1pt}},
  after upper={\vspace{1pt}},
  parskip=0pt
    ]
    
    All scores are Likert scores, integers between 1--5.
    
    \subsubsection*{Instruction Quality and Diversity}
    We wish to measure quality and diversity of the generated instructions by manually rating the Naturalness (for quality), and specificity (for diversity).
    
    \begin{enumerate}[topsep=0pt,itemsep=-1ex,partopsep=1ex,parsep=1ex]
    \item \textbf{Naturalness} -- The instruction is clear, fluent, plausible, and relevant to the context of the document set and its topic.
      \begin{itemize}[topsep=0pt,itemsep=-1ex,partopsep=1ex,parsep=1ex]
        \item \textbf{Clear} -- The instruction is unambiguous and understandable, given the topic.
        \item \textbf{Fluent} -- The instruction is written in natural, human-like language and adheres to proper grammar.
        \item \textbf{Plausible} -- The instruction appears to be written by a human and does not resemble machine-generated instructions.
        \item \textbf{Contextually Relevant} -- The instruction aligns with the context of the document set and reflects a use case that a human might reasonably formulate.
          \begin{itemize}[topsep=0pt,itemsep=-1ex,partopsep=1ex,parsep=1ex]
            \item Based on general knowledge or expertise (e.g., as an NLP student), the use case appears likely.
          \end{itemize}
        \item Example of low naturalness -- \textit{``Select content that lists the factors increasing the likelihood of resumption of ovarian cyclicity (ROC) at \uline{36 to 42 days} in milk (DIM) from \uline{Document \#1}''}.
      \end{itemize}
    
    \item \textbf{Specificity} -- The information sought by the instruction is central (1) or anecdotal (5) to the topic of the document set.
      \begin{itemize}[topsep=0pt,itemsep=-1ex,partopsep=1ex,parsep=1ex]
        \item To assess specificity, imagine a conceptual tree of all information organized hierarchically within the document set. The measure reflects how deep into the tree the instruction seeks information.
        \item \textbf{Levels:}
          \begin{enumerate}[topsep=0pt,itemsep=-1ex,partopsep=1ex,parsep=1ex]
            \item \textbf{Topic Level} -- The instruction addresses the broad topic of the document set.
            \item \textbf{Sub-Topic Level} -- The instruction focuses on a specific sub-topic within the broader topic.
            \item \textbf{Somewhat Central} -- The instruction targets a detail that is central but not overly specific.
            \item \textbf{Anecdotal Point} -- The instruction seeks information tied to a specific detail or anecdote.
            \item \textbf{Atomic Anecdotal Point} -- The instruction zeroes in on a singular, highly specific detail related to the topic.
          \end{enumerate}
      \end{itemize}
    \end{enumerate}

    \subsubsection*{Content Selection}
    Select the relevant text spans from the document set based on the given instruction.

    \end{tcolorbox}
    \caption{Manual annotation guidelines for \textit{naturalness} and \textit{specificity} for rating \ri{} instructions.}
\label{fig:annotation-guidelines}
\end{figure*}

\paragraph{Inter-rater agreement of selections.}
\label{app:ri-qa-iaa}

To quantify the agreement, we cannot directly use Cohen's $\kappa$ as it requires the number of negative cases, which is ill-defined (or very large) in the content selection setting (e.g. all the spans that are not part of the selection).
Instead, we followed \citet{hripcsak_agreement_2005}, who demonstrated that as the number of possible negative cases increases, $\kappa$ approaches the average $F_1$ score, which is the case in our content selection setting.
Based on this, to compute $\kappa$ between two groups of annotators, we computed the average pairwise token-level $F_1$ score (\S\ref{subsection:evaluation}) between every possible pair of annotators, where each annotator originates from a different group.

\begin{table}[htb]
\resizebox{\columnwidth}{!}{%
\begin{tabular}{@{}llll@{}}
\toprule
Task & Pearson $r$ & Spearman $\rho$ & Kendall $\tau$ \\ \midrule
\scifactshort{}        & 0.966    & 0.955    & 0.83          \\
\openaspshort{}        & 0.967    & 0.976     & 0.886          \\
\aspectnewsshort{}     & 0.867   & 0.85    & 0.697          \\
\debatesumshort{}      & 1.0     & 0.999    & 0.993          \\ \bottomrule
\end{tabular}%
}
\caption{
System-level correlations between the original evaluation metrics used in  four of the benchmark tasks and the proposed token-level $F_1$ metric (Section~\ref{subsection:evaluation}), which we suggest as a standardized metric for IGCS tasks. Overall, the token-level $F_1$ demonstrates strong to very strong correlation with all other metrics, indicating its reliability for evaluating content selection performance.
}
\label{tab:meta-eval-results}
\end{table}

\section{Model Configurations}
\label{app:models-training-details}

We trained Llama-3-8B \texttt{meta-llama/ Meta-Llama-3-8B-Instruct} with various data mixtures, as described in \S\ref{subsec:ft-models}.
Each dataset training mix was shuffled before training.

For training, we first experimented with several hyperparameter options. We also tuned the prompt manually to make it work better for the zero-shot case. For models trained on multiple training sets, we attempted to balance the different sets by up-sampling smaller datasets to match the size of the largest dataset in the mix. We found this approach to be inferior and as such abandoned this technique.

The input prompt is shown in \autoref{fig:app_prompt_igcs}. The target output is a JSON array of the selected texts.

All tested models accommodated the input size except for two \openaspshort{} instances on Llama-3-8B and Llama-3-70B without document-level inference. In those cases, we truncated the input to fit the 8K token context size.

All model variants were trained on three A100 GPUs with a maximum sequence length of 4096, batch size of 4, NEFTune noise $\alpha=5.0$ \cite{jain2023neftunenoisyembeddingsimprove}, and a warmup ratio of 0.06 for 3 epochs.
Training each model variant typically took several hours on the next token prediction task.
We trained each model on the next-token completion task, ignoring system and user prompts in the loss function.


During the decoding inference phase, all models (GPT-4, Claude-3-Opus, Gemini-1.5-Pro, and the three versions of Llama-3) were set to a temperature of 0.0 to ensure reproducibility.
We set \texttt{max\_new\_tokens} to 2048 for Llama-3-8B and its trained variants.

For the Llama-3 70B and 405B models, we used Together AI's API\footnote{\url{https://www.together.ai}} with the model versions \texttt{meta-llama/Llama-3-70b-chat-hf} and \texttt{meta-llama/Meta-Llama-3.1-405B-
Instruct-Turbo}, respectively.

For the Qwen 2.5 family, we used the instruct variants (e.g., \texttt{Qwen/Qwen2.5-7B-Instruct}) of the four smallest models --- 0.5B, 1.5B, 3B, and 7B. Similarly, for the SmolLM2 family, we used the instruct variants (e.g., \texttt{HuggingFaceTB/SmolLM2-1.7B-\\Instruct}) of all three models --- 135M, 360M, and 1.7B.

\section{Fuzzy Match}
\label{app:models-grounding}

We describe here the fuzzy match grounding algorithm, mentioned in \S\ref{sec:model}, which grounds the output text to a location in the source text.
Since LLMs sometimes paraphrase or slightly alter copied text spans, we relax the exact (case-insensitive) textual search.
We consider the output text to be matched to a (non-empty) sub-sequence in the source text when there is a token-level Levenshtein distance of up to 15\% the length of the output (and no more than 10 tokens) between the two strings.
If there are multiple matches, we select the one with the lowest edit distance that appears first.
When no such match is found, the text span is discarded from the suggested selection.
We tested larger Levenshtein distances, which result in a rapidly growing computational overhead, but have minimal or no positive effect.

\def\aspectnewsPOne{Select between 1 and 3 sentences from the provided news article that are most relevant to the \{topic\}'s \{aspect\_description\}.

Read the news article carefully and identify all sentences. Internally analyze each sentence for relevance to the specified topic and aspect. Perform your reasoning steps before arriving at a final conclusion, but only output the final result. Do not modify or alter any of the selected sentences.

\# Steps
1. **Parse the Article:** Break the article into individual sentences.
2. **Internal Reasoning:** Evaluate each sentence for its relevance to the \{topic\}'s \{aspect\_description\} based on the context and details provided.
3. **Selection:** Choose at least 1 and at most 3 sentences that best capture the required information.
4. **Conclusion:** Prepare your final selection after completing your internal reasoning.

\# Output Format
Output a valid JSON array of strings containing the exact selected sentence(s). For example: ["Sentence 1", "Sentence 2"]

\# Notes
- Do not change or rephrase any of the copied sentence(s).
- Ensure that your internal reasoning process is used to determine the selection, but do not include it in the final output.}
\def\aspectnewsPTwo{From the news article below, pick between 1 and 3 sentences that best address the \{topic\}’s \{aspect\_description\}. Return them verbatim as a valid JSON array of strings.}
\def\aspectnewsPOneICL{Given the following news article, select at least 1 and at most 3 sentences that are the most relevant to the given aspect. Output the exact sentences from the given document as a valid json array of strings. Do not change the copied text.
Below \{is an example | are examples\} of an input and the correct selected content:

Aspect: \{example1\_topic\}'s \{example1\_aspect\_description\}

Input Document(s):

\{example1\_documents\}

...

--- END OF EXAMPLES ---

Now, select content from the below document(s):

Aspect: \{topic\}'s \{aspect\_description\}

Input Document(s):

\{documents\}
}

\def\aspectnewsPTwoICL{From the news article below, pick between 1 and 3 sentences that best address the aspect. Return them verbatim as a valid JSON array of strings.

Below \{is an example | are examples\} of an input and the correct selected content:

Aspect: \{example1\_topic\}'s \{example1\_aspect\_description\}

Input Document(s):

\{example1\_documents\}

...

--- END OF EXAMPLES ---

Now, select content from the below document(s):

Aspect: \{topic\}'s \{aspect\_description\}

Input Document(s):

\{documents\}
}
\def\debatesumPOne{Summarize Evidence for Argument

Task:  
Given a document and a specific argument represented by "\{argument\}", scan the document and identify short, concise text phrases that serve as evidence supporting that argument. Your task is to extract these evidence phrases exactly as they appear in the document without any modifications.

Additional Details:  
- Examine the entire document carefully to locate all relevant segments that directly support the argument.  
- Do not paraphrase or change any portion of the selected text.  
- Your response must include a reasoning section that details how you identified and selected the evidence phrases, followed by a conclusion section with the final output.

Steps:  
1. **Reasoning:**  
   - Read through the provided document carefully.  
   - Identify text segments that clearly support the argument "\{argument\}".  
   - Document your thought process detailing how you determined which phrases were relevant.  
   - **Note:** The reasoning section must come first in your response.

2. **Conclusion:**  
   - List the selected evidence phrases exactly as they appear in the document.  
   - Format your answer as a valid JSON array of strings.  
   - Ensure that the JSON output contains no additional text or modifications to the evidence.

Output Format:  
- The response must have two clearly separated sections:  
  - **Reasoning:** A detailed explanation of the extraction process.  
  - **Conclusion:** A JSON array (e.g., ["Phrase one", "Phrase two", "..."]) containing the exact evidence phrases from the document.

Notes:  
- The reasoning process must precede the final conclusion.  
- Do not alter, reword, or modify any of the text phrases extracted from the document.  
- Ensure that the final JSON output is syntactically valid and consists solely of the evidence phrases.}
\def\debatesumPTwo{Given the document below, extract short, concise excerpts that cover every piece of evidence for the argument "\{argument\}". Return those exact excerpts—unaltered—as a JSON array of strings.}
\def\debatesumPOneICL{Given the following document, select short and concise text phrases that summarize all the evidence for the given argument. Output the exact text phrases from the given document as a valid json array of strings. Do not change the copied text.
Argument: \{argument\}}
\def\debatesumPTwoICL{Given the following document, select short and concise text phrases that summarize all the evidence for the given argument. 
Argument: \{argument\}}
\def\scifactPOne{Extract evidence sentences from provided medical abstracts for the claim.

Additional details:
- Input includes a claim in the form "\{claim\}" and one or more abstracts of medical papers.
- Your objective is to identify and extract any sentences that directly provide supporting or refuting evidence for the given claim.
- Do not modify, paraphrase, or change the exact text of any sentence extracted from the abstracts.
- Ensure that you examine each sentence methodically and conduct a reasoning process before finalizing the result. The reasoning steps must come before any conclusion, and the final output should present the selected sentences after all reasoning is complete.

Steps:
1. Parse the input to obtain the claim and the provided abstract document(s).
2. Split each abstract into individual sentences.
3. Evaluate each sentence to determine if it offers either supporting or refuting evidence for the claim.
4. Ensure all extracted sentences are unaltered in punctuation, casing, and formatting.
5. Reason through the evidence identification process before arriving at the conclusion.

Output Format:
- Provide your final result as a valid JSON array of strings.
- The JSON array must contain only the exact sentences that offer evidence for the claim.
- No extra commentary, reasoning details, or additional text should be included in the final output.

Notes:
- Ensure that the reasoning process is clearly organized with the reasoning steps listed before the final answer, but only the final JSON array should be output as the conclusion.
- Refrain from including any explanations or reasoning within the final output.
- Follow all instructions closely, ensuring minimal changes to the original text while preserving clarity and precision in the final result.}
\def\scifactPTwo{From the provided medical abstract(s), identify sentences that support or oppose the claim: "\{claim\}". Return those sentences verbatim as a JSON array of strings.}
\def\scifactPOneICL{Given the following abstract document(s) of medical papers, select the sentences that provide either supporting or refuting evidence for the given claim". Output the exact sentences from the given abstract(s) as a valid json array of strings. Do not change the copied text.
Claim: \{claim\}}
\def\scifactPTwoICL{From the provided medical abstract(s), identify sentences that support or oppose the claim. Return those sentences verbatim as a JSON array of strings.
Claim: \{claim\}}
\def\openaspPOne{Extract all sentences from the provided news articles that directly relate to the aspect specified by "\{aspect\_label\}" while preserving the exact original wording.

Ensure you follow these details:
- Read each news article thoroughly for content related to "\{aspect\_label\}".
- Identify and select sentences that explicitly mention or contextually relate to the specified aspect.
- **Do not rephrase, summarize, or modify any text.** Each extracted sentence must be copied exactly as it appears in the source.
- Begin by outlining your reasoning process—describe your criteria and steps for determining relevance—before presenting your final output. The reasoning steps must occur first, and the extracted sentences (final conclusions) should appear last.

\# Steps
1. Analyze each provided news article sentence by sentence.
2. Determine which sentences are relevant to "\{aspect\_label\}" based on explicit mentions or contextual clues.
3. Record your reasoning process briefly (i.e., criteria for selection and decision points) before compiling the final results.
4. Compile all relevant sentences into a list.

\# Output Format
Provide your final answer as a valid JSON array of strings. For example:  
["First relevant sentence.", "Second relevant sentence.", "Third relevant sentence."]

\# Notes
- The reasoning process must precede the final output and should clearly describe the selection criteria and process.
- Do not alter any of the extracted sentences in any way; maintain the original text in full.
- The final JSON array must be strictly valid without additional commentary or formatting outside of it.}
\def\openaspPTwo{From the provided news articles on the topic "\{title\}", extract every sentence pertaining to "\{aspect\_label\}" and return them verbatim as a JSON array of strings. Do not modify the text.}
\def\openaspPOneICL{Given the following news articles on the same topic, extract all sentences related to the given aspect. Output the exact sentences from the given documents as a valid json array of strings. Do not change the copied text.
Title: \{title\}
Aspect: \{aspect\_label\}}
\def\openaspPTwoICL{From the provided news articles on the topic, extract every sentence pertaining to the aspect and return them verbatim as a JSON array of strings. Do not modify the text.
Title: \{title\}
Aspect: \{aspect\_label\}}
\def\evidencedetectionPOne{Extract short and concise text phrases from provided documents that reference the query.

Additional details:
- You are given multiple documents on the same topic and a specific query placeholder "\{query\}".
- Your goal is to extract the exact text phrases from these documents that support or provide references to the query.
- Do not alter or modify any copied text; use the text exactly as it appears in the documents.

\# Steps
1. Read and comprehend the documents and the query.
2. Identify text segments in the documents that reference or support the query "\{query\}".
3. Detail your reasoning steps by explaining how you determined which phrases were relevant.
4. Validate that the selected text phrases are exact matches from the source documents.
5. Structure your output so that the detailed reasoning is presented first, followed by the final JSON output.

\# Output Format
- Provide the reasoning process in clear, logical steps.
- Conclude with a final answer as a valid JSON array of strings.
- Do not modify or paraphrase any extracted text; include the copied text exactly as in the documents.

Example Output:  
Reasoning: [Explain step-by-step how the relevant text phrases were identified from the documents, ensuring that reasoning appears before the conclusion.]  
Final Answer: ["Extracted text phrase 1", "Extracted text phrase 2"]

\# Notes
- Ensure reasoning always appears before any conclusions or final outputs.
- The final output must strictly be a JSON array of strings containing the extracted phrases.
- If no relevant phrases are found in a document, omit that document from your output.}
\def\evidencedetectionPTwo{Given the following documents on the same topic, extract brief, precise text snippets that support the statement "\{query\}". Output these exact snippets verbatim as a JSON array of strings.}
\def\evidencedetectionPOneICL{Given the following documents on the same topic, extract short and concise text phrases that provide references to the given statement. Output the exact text phrases from the given documents as a valid json array of strings. Do not change the copied text.
Statement: \{query\}}
\def\evidencedetectionPTwoICL{Given the following documents on the same topic, extract brief, precise text snippets that support the statement. Output these exact snippets verbatim as a JSON array of strings.
Statement: \{query\}}
\def\saliencydetectionPOne{Extract short and concise salient text phrases from the provided documents.

Read the provided documents on the same topic and extract text segments that clearly represent the key points. Only extract text phrases that are exactly present in the original documents without any modifications. The phrases must be short, concise, and capture the essential information. Do not rephrase or alter the original text in any way.

**Reasoning Steps:**
1. Read and analyze each provided document carefully.
2. Identify text segments that are short, clear, and capture the most salient points.
3. Verify that each extracted phrase appears exactly as in the documents.
4. Perform all reasoning steps before concluding with the final output.

**Output Format:**
- The final answer must be a valid JSON array of strings.
- Each string in the array should be a salient text phrase extracted from the documents.

**Notes:**
- Follow the structured reasoning steps before listing the final result.
- Do not modify, rephrase, or change the original text of any extracted phrase.
- Reasoning must precede conclusions, and final output should be the extracted phrases only, formatted as specified.}
\def\saliencydetectionPTwo{Given the following documents on the same topic, extract short, salient phrases exactly as written. Return them as a valid JSON array of strings. Do not alter the copied text.}
\def\saliencydetectionPOneICL{Given the following documents on the same topic, extract short and concise salient text phrases. Output the exact text phrases from the given documents as a valid json array of strings. Do not change the copied text.
}
\def\saliencydetectionPTwoICL{Given the following documents on the same topic, extract short, salient phrases exactly as written. Return them as a valid JSON array of strings. Do not alter the copied text.
}

\begin{table*}[t]
  \centering
  \vsmall
  \begin{tabularx}{\textwidth}{@{}X@{}}
    \toprule

    \textbf{\openaspshort{}'s Zero-shot Prompt Variant 1} \\ \hline
    \aspectnewsPOne{} \\  
    \addlinespace

    \textbf{\openaspshort{}'s Zero-shot Learning Prompt Variant 2} \\ \hline
    \aspectnewsPTwo{} \\
    \addlinespace

    \textbf{\openaspshort{}'s In-context Learning Default Prompt} \\ \hline
    \aspectnewsPOneICL{} \\
    \addlinespace

    \textbf{\openaspshort{}'s In-context Learning Prompt Variant 1} \\ \hline
    \aspectnewsPTwoICL{} \\
    \addlinespace

    \bottomrule
  \end{tabularx}
  \caption{Zero-shot and In-context learning prompt variants for \openaspshort{}. The default zero-shot prompt template can be found in \autoref{fig:app_prompt_igcs}.}
  \label{tab:prompt-variants}
\end{table*}

\section{Correlation of Evaluation Metrics - Complementary Results}
\label{app:meta-eval}

In \autoref{tab:meta-eval-results}, we report the system-level correlation coefficients between each task-specific metric and the generic token-level metric (\S\ref{subsection:evaluation}).
These coefficients are computed using system-level scores from 24 model configurations for \debatesumshort{} and \aspectnewsshort{}, and from 35 configurations for \openaspshort{} and \scifactshort{}.
The configurations vary along three dimensions: the choice of fine-tuning training set, the use or omission of in-context learning, and the application of document-level inference (the latter applicable only to \openaspshort{} and \scifactshort{}).

\section{Transfer Learning - Complementary Results}
\label{app:detailed-tl-results}

In \autoref{tab:app-main-results-detailed} we show the original token-level $F_1$ scores for the six \igcsbench{} tasks.

\section{Significance Testing}\label{app:significance-testing}

As some task-specific metrics are defined at the system level, we used a permutation test to measure the significance (at $p < 0.05$) of the difference between two model scores, performing random sampling with a size of 1,000 for each test \cite{noreen1989computer}.

\begin{table*}[tb]
\resizebox{\textwidth}{!}{%
\begin{tabular}{@{}clrrrrrrrrcc@{}}
\toprule
\multicolumn{1}{l}{}                                                                                                                & Models              & \multicolumn{2}{c}{\aspectnewsshort{}}         & \multicolumn{2}{c}{\debatesumshort{}}          & \multicolumn{2}{c}{\scifactshort{}}            & \multicolumn{2}{c}{\openaspshort{}}            & \multicolumn{1}{c}{\saliencyshort{}} & \multicolumn{1}{c}{\evidencedetectionshort{}} \\
\multicolumn{1}{l}{}                                                                                                                &                     & \multicolumn{1}{c}{F1} & \multicolumn{1}{c}{O} & \multicolumn{1}{c}{F1} & \multicolumn{1}{c}{O} & \multicolumn{1}{c}{F1} & \multicolumn{1}{c}{O} & \multicolumn{1}{c}{F1} & \multicolumn{1}{c}{O} & \multicolumn{1}{c}{F1/O}             & \multicolumn{1}{c}{F1/O}                      \\ \midrule
\multicolumn{1}{l}{}                                                                                                                & \llamaIcl{}         & 59.2                   & 34.6                  & 46.6                   & 46.9                  & 58.0                   & 44.8                  & 27.1                   & 28.6                  & 42.9                                 & 13.5                                          \\
\multicolumn{1}{l}{}                                                                                                                & \llamaVanilla{}     & 56.1                   & 29.4                  & 42.4                   & 42.4                  & 46.6                   & 47.5                  & 42.3                   & 41.9                  & 36.6                                 & 27.3                                          \\   \cdashline{2-12}
\multirow{-2}{*}{\textbf{\makecell[cc]{\textbf{\rotatebox[origin=c]{90}{transfer-}} \textbf{\rotatebox[origin=c]{90}{only}}}}} & + \loo{}            & 41.6                   & 34.0                  & 29.3                   & 29.1                  & 23.8                   & 25.8                  & 32.0                   & 30.4                  & 41.9                                 & 10.2                                          \\
                                                                                                                                    & + \igcsriunion{}    & 63.3                   & 37.0                  & 36.6                   & 36.7                  & 51.3                   & 42.6                  & 52.0                   & 49.3                  & 37.5                                 & 33.6                                          \\
                                                                                                                                    & + \igcsrimajority{} & 57.6                   & 35.6                  & 25.2                   & 25.2                  & 58.9                   & 48.1                  & 49.7                   & 47.1                  & 32.4                                 & 35.6                                          \\ \midrule
\multirow{1}{*}{\textbf{\makecell[cl]{\textbf{\rotatebox[origin=c]{90}{\footnotesize{supervision+}}} \textbf{\rotatebox[origin=c]{90}{\footnotesize{transfer}}}}}} & \llamaSup{}         & 69.5                   & 40.6                  & 64.7                   & 63.5                  & 74.4                   & 66.0                  & --                     & --                    & --                                   & --                                            \\   \cdashline{2-12}
                                                                                                                                    & + \loo{}            & 72.4                   & 42.3                  & 65.3                   & 64.1                  & 78.9                   & 70.0                  & --                     & --                    & --                                   & --                                            \\
                                                                                                                                    & + \igcsriunion{}    & 72.7                   & 42.7                  & 64.9                   & 63.7                  & 81.1                   & 72.1                  & --                     & --                    & --                                   & --                                            \\
                                                                                                                                    & + \igcsrimajority{} & 75.4                   & 43.2                  & 64.7                   & 63.6                  & 79.8                   & 68.8                  & --                     & --                    & --                                   & --                                            \\ \midrule
\multirow{5}{*}{\textbf{\makecell[cl]{\textbf{\rotatebox[origin=c]{90}{prompt-based}} \textbf{\rotatebox[origin=c]{90}{models}}}}} & \llamaM{}           & 50.5                   & 29.3                  & 41.1                   & 40.7                  & 65.2                   & 58.5                  & 51.9                   & 56.8                  & 33.2                                 & 44.9                                          \\
                                                                                                                                    & \llamaL{}           & 53.7                   & 30.0                  & 45.5                   & 45.4                  & 61.8                   & 56.2                  & 57.8                   & 59.8                  & 35.1                                 & 42.1                                          \\
                                                                                                                                    & \gpt4{}             & 60.4                   & 32.8                  & 39.5                   & 39.0                  & 63.6                   & 58.6                  & 55.4                   & 57.4                  & 39.1                                 & 50.1                                          \\
                                                                                                                                    & \gptIcl{}           & 63.8                   & 33.9                  & 46.0                   & 45.5                  & 63.2                   & 57.3                  & 52.9                   & 55.0                  & 39.9                                 & 47.5                                          \\
                                                                                                                                    & \claude{}           & 51.7                   & 31.3                  & 50.0                   & 49.6                  & 57.2                   & 54.5                  & 53.9                   & 52.2                  & 43.7                                 & 28.2                                          \\ \bottomrule
\end{tabular}%
}
\caption{Comparison between our proposed token-level \textbf{$F_1$} (\S\ref{subsection:evaluation}) and the original metric defined for each task, denoted as \textbf{O}. For \saliencyshort{} and \evidencedetectionshort{} the original metric is the same. Model configurations and fine-tuning details are described in \S\ref{sec:model} and follow the same notations as \autoref{tab:main-results}.}
\label{tab:app-main-results-detailed}
\end{table*}

\section{Document-level Inference - Complementary Results}
\label{app:detailed-frag-results}

\begin{table*}[t]
\resizebox{\textwidth}{!}{%
\begin{tabular}{@{}lrrrrrrrrrrrr@{}}
\toprule
Model                & \multicolumn{3}{c}{\evidencedetectionshort{}}                                                        & \multicolumn{3}{c}{\scifactshort{}} &       \multicolumn{3}{c}{\saliencyshort{}}      & \multicolumn{3}{c}{\openaspshort{}}               \\
                     & \multicolumn{1}{c}{Score} & \multicolumn{1}{c}{$\Delta$} & \multicolumn{1}{c}{$\Delta$ Sel.} & \multicolumn{1}{c}{Score}           & \multicolumn{1}{c}{$\Delta$} & \multicolumn{1}{c}{$\Delta$  Sel.} & \multicolumn{1}{c}{Score}            & \multicolumn{1}{c}{$\Delta$} & \multicolumn{1}{c}{$\Delta$ Sel.} & \multicolumn{1}{c}{Score}           & \multicolumn{1}{c}{$\Delta$} & \multicolumn{1}{c}{$\Delta$ Sel.} \\ \midrule
\llama{}             & 29.6                      & -2.4                               & +64.9                               & 44.8                                & +2.7                               & +7.2                                & 31.7                                 & +4.9                               & +142                                & 18.4                                & +23.4                              & +741                                \\
\llamaM{}            & 42.8                      & +2.1                               & +12.7                               & 57.5                                & +1.0                               & +0.8                                & 28.9                                 & +4.3                               & +126                                & 24.6                                & +32.2                              & +478                                \\
\llamaL{}            & 40.4                      & +1.6                               & +34.8                               & 56.7                                & -0.5                               & +1.0                                & 31.7                                 & +3.4                               & +192                                & 43.8                                & +16.0                              & +454                                \\
\llamaRiUnion{}      & 32.1                      & +1.5                               & +21.9                               & 42.0                                & +0.6                               & +4.3                                & 31.0                                 & +6.6                               & +282                                & 29.5                                & +19.8                              & +643                                \\
\llamaRiMajority{}   & 33.6                      & +2.0                               & +21.3                               & 47.8                                & +0.2                               & +3.3                                & 21.3                                 & +11.0                              & +146                                & 18.7                                & +28.4                              & +578                                \\
\llamaFullUnion{}    & 27.9                      & -8.3                               & +213.7                              & 70.6                                & +0.2                               & +0.8                                & 37.9                                 & +3.4                               & +260                                & 20.9                                & +21.9                              & +992                                \\
\llamaFullMajority{} & 25.9                      & -10.6                              & +289.5                              & 69.6                                & +0.5                               & +0.1                                & 32.6                                 & +8.4                               & +352                                & 21.8                                & +23.3                              & +795                                \\
\gpt4{}              & 51.7                      & -1.6                               & +21.5                               & 59.1                                & -0.5                               & +2.4                                & 32.9                                 & +6.2                               & +246                                & 39.3                                & +18.1                              & +485                                \\
\claude{}            & 36.8                      & -8.6                               & +79.7                               & 53.7                                & +0.8                               & +7.1                                & 36.6                                 & +7.1                               & +247                                & 35.9                                & +16.3                              & +564                                \\ \midrule
Average              & \multicolumn{1}{l}{}      & -2.7                               & +84.4                               & \multicolumn{1}{l}{}                & +0.6                               & +3.0                                & \multicolumn{1}{l}{}                 & +6.1                               & +222                                & \multicolumn{1}{l}{}                & +22.2                              & +636                                \\ \bottomrule
\end{tabular}%
}
\caption{The performance of the four multi-document tasks, detailed in \S\ref{subsec:results-inference-config}, is presented as task-specific metric scores.  
\textbf{Score} refers to the task score without applying document-level inference (\S\ref{subsec:model-frag}); \textbf{$\Delta$} indicates the performance gain when employing the document-level inference method; \textbf{$\Delta$ Sel.} indicates the difference in the size of the model's output, measured as the average number of tokens in the selection.
The last row shows the average difference across all models for each task.}
\label{tab:app-detailed-frag}
\end{table*}

In \autoref{tab:app-detailed-frag}, we show the scores of nine models tested on the four multi-document tasks in \igcsbench{}, from which \autoref{fig:bar_fragment_diff} is derived.

\section{Prompt Robustness Results}
\label{app:prompt-robustness}
\begin{figure*}[t]
    \centering
    \includegraphics[width=1.0\textwidth]{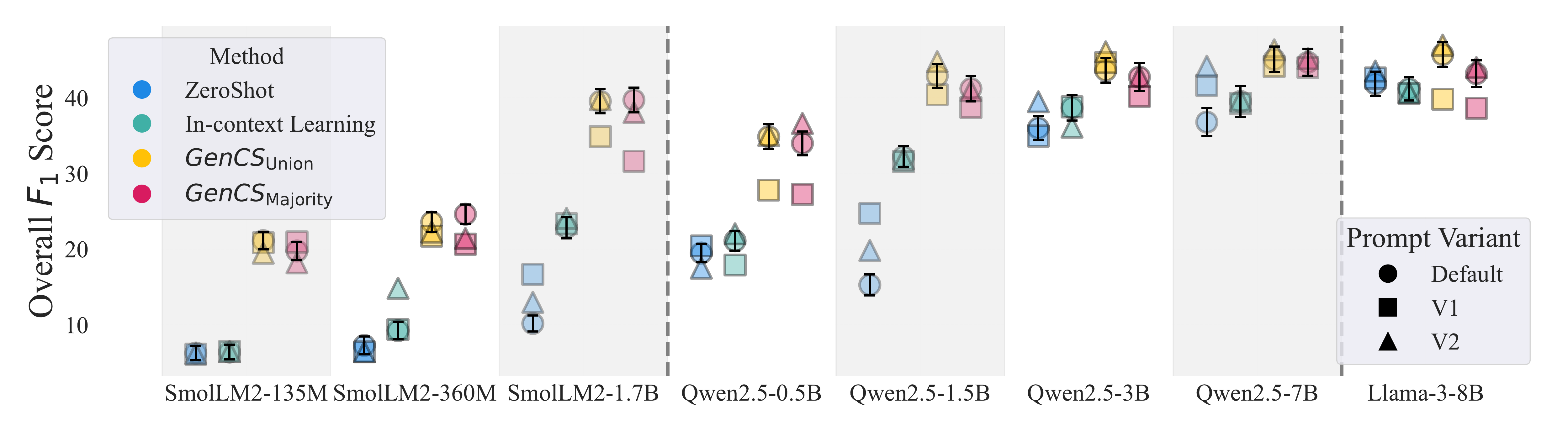}
    \caption{
    Overall $F_1$ scores on \igcsbench{} for eight models, evaluated across four methods and three prompt variants (see examples for the prompt variants in \autoref{tab:prompt-variants}). For comparison, see \autoref{fig:bars_overall_scores}, which presents results using only the default prompt. Confidence intervals ($\alpha=0.05$) are shown for the default prompt. While performance varies across prompt variants, the overall trends remain consistent with those reported in Section~\ref{subsec:results-main} --- most models benefit from \ri{} fine-tuning, with the smallest models exhibiting the largest gains.
    }
    \label{fig:prompt_robustness}
\end{figure*}

To further validate our findings in the transfer-only setting (discussed in \S\ref{par:results-main-zeroshot} and exhibited in \autoref{fig:bars_overall_scores}), we conducted each experiment with two additional prompt variants automatically tuned via meta-prompting using OpenAI's O4-mini-high.\footnote{\url{https://openai.com/index/introducing-o3-and-o4-mini}}
The resulting prompt variants and the default in-context learning prompt are exemplified for a single task in \autoref{tab:prompt-variants}.
Notably, these variants are not mere paraphrases of the original instructions but substantially different in terms of length and details for each of the six tasks. Thus they provide more variability for the analysis.
For the in-context learning scenario, one of the prompt variants (V1) used a one-shot example instead of two-shot, to additionally explore different number of samples.
The fine-tuned models, namely \igcsriunion{} and \igcsrimajority{}, were fine-tuned using the default prompt, suggesting adaptability of the fine-tuned models to prompt variations.

In \autoref{fig:prompt_robustness}, we observe trends similar to those in \autoref{fig:bars_overall_scores}, with significant performance gains across most models under different prompt variants. 
Moreover, we note variability in model performance across these prompts, a phenomenon reported in prior works \cite{DBLP:conf/iclr/Sclar0TS24, mizrahi-etal-2024-state}.

\end{document}